%% file: main.tex
\documentclass[11pt,a4paper]{article}
\usepackage{acl}
\usepackage{arydshln}
\usepackage{booktabs}
\usepackage{times}
\usepackage{amssymb}
\usepackage{xcolor}
\usepackage{forest}
\usepackage{graphicx}
\usepackage{float}
\usepackage{tikz-qtree}
\usepackage{latexsym}
\usepackage{xspace}
\usepackage{algorithm}
\usepackage{algorithmicx}
\usepackage{inconsolata}
\usepackage[title]{appendix}

\usepackage{enumitem,kantlipsum}
\usepackage{amsmath}

\usepackage{microtype}

\def\SPSB#1#2{\rlap{\textsuperscript{\textcolor{blue}{#1}}}\SB{#2}}

\def\SB#1{\textsubscript{\textcolor{blue}{#1}}}
\def\SBR#1{\textsubscript{\textcolor{red}{#1}}}

\newcommand{\exa}[1]{``{#1}''}

\newcommand{\bl}[1]{\textcolor{blue}{#1}}

\usepackage[position=b]{subcaption}
\usepackage{graphicx}
\usepackage{hyperref}
\usepackage{linguex}
\usepackage{tikz}
\usepackage{fixmetodonotes}

\title{The Paradox of the Compositionality of Natural Language: \\A Neural Machine Translation Case Study}

\author{Verna Dankers \\
  ILCC, University of Edinburgh \\
  \texttt{\small vernadankers@gmail.com} \\\And
  Elia Bruni \\
  University of Osnabr\"uck \\
  \texttt{\small elia.bruni@gmail.com} \\\And
  Dieuwke Hupkes \\
  Facebook AI Research \\
  \texttt{\small dieuwkehupkes@fb.com}}

\begin{document}
\maketitle

\input{abstract}

\input{introduction}

\input{tables/template_table}

\input{background}

\input{setup}

\input{tests}

\input{manual_analysis}

\input{related_work}

\input{discussion}

\input{acknowledgments}

\bibliography{references,anthology}
\bibliographystyle{acl_natbib}

\clearpage

\appendix
\begin{appendices}
\input{appendix}
\input{appendix_analysis}
\end{appendices}

\end{document}

%% file: abstract.tex
\begin{abstract}
Obtaining human-like performance in NLP is often argued to require compositional generalisation. 
Whether neural networks exhibit this ability is usually studied by training models on highly compositional synthetic data.
However, compositionality in natural language is much more complex than the rigid, arithmetic-like version such data adheres to, and artificial compositionality tests thus do not allow us to determine how neural models deal with more realistic forms of compositionality.
In this work, we re-instantiate three compositionality tests from the literature and reformulate them for \emph{neural machine translation} (NMT). 
Our results highlight that: 
    i) unfavourably, models trained on \emph{more} data are more compositional;
    ii) models are sometimes less compositional than expected, but sometimes more, exemplifying that different \emph{levels} of compositionality are required, and models are not always able to modulate between them correctly;
    iii) some of the non-compositional behaviours are mistakes, whereas others reflect the natural variation in data.
Apart from an empirical study, our work is a call to action:
we should rethink the evaluation of compositionality in neural networks and develop benchmarks using \emph{real} data to evaluate compositionality on natural language, where composing meaning is not as straightforward as doing the math.\footnote{The data and code are available at \url{https://github.com/i-machine-think/compositionality_paradox_mt}. We present details concerning reproducibility in Appendix~\ref{app:reproducibility}.}
\end{abstract}

%% file: introduction.tex
\section{Introduction}

Although the successes of deep neural networks in \textit{natural language processing} (NLP) are astounding and undeniable, they are still regularly criticised for lacking the powerful generalisation capacities that characterise human intelligence.
A frequently mentioned concept in such critiques is \emph{compositionality}: the ability to build up the meaning of a complex expression by combining the meanings of its parts \citep[e.g.][]{partee1984compositionality}.
Compositionality is assumed to play an essential role in how humans understand language, but whether neural networks also exhibit this property has since long been a topic of vivid debate \citep[e.g.][]{fodor1988connectionism,smolensky1990tensor,marcus2003algebraic,nefdt2020puzzle}.

Studies about the compositional abilities of neural networks consider almost exclusively models trained on synthetic datasets, in which compositionality can be ensured and isolated \citep[e.g.][]{lake2018generalization,hupkes2020compositionality}.\footnote{
Apart from \citet{raunak2019compositionality}, work on compositionality and `natural' language considers highly structured subsets of language \citep[e.g.][]{kim2020cogs,keysers2019measuring}.}
In such tests, the interpretation of expressions is computed completely \emph{locally}: every subpart is evaluated independently -- without taking into account any external context -- and the meaning of the whole expression is then formed by combining the meanings of its parts in a bottom-up fashion.
This protocol matches the type of compositionality observed in arithmetic: the meaning of $(3 + 5)$ is always $8$, independent of the context it occurs in.

However, as exemplified by the sub-par performance of symbolic models that allow only strict, local protocols, compositionality in natural domains is far more intricate than this rigid, arithmetic-like variant of compositionality.
Natural language seems very compositional, but at the same time, it is riddled with cases that are difficult to interpret with a strictly local interpretation of compositionality.
Sometimes, the meaning of an expression does not derive from its parts (e.g. for idioms), but the parts themselves are used compositionally in other contexts.
In other cases, the meaning of an expression does depend on its parts in a compositional way, but arriving at this meaning requires a more \textit{global} approach because the meanings of the parts need to be disambiguated by information from elsewhere.
For instance, consider the meaning of homonyms (``these dates are perfect for our dish/wedding''), potentially idiomatic expressions (``the child kicked the bucket off the pavement''), or scope ambiguities (``every human likes a cat'').
This paradoxical tension between local and global forms of compositionality inspired many debates on the compositionality of natural language.
Likewise, it impacts the evaluation of compositionality in NLP models.
On the one hand, local compositionality seems necessary for robust and reliable generalisation.
Yet, at the same time, global compositionality is needed to appropriately address the full complexity of language, which makes evaluating compositionality of state-of-the-art models `in the wild' a complicated endeavour.

In this work, we face this challenge head-on.
We concentrate on the domain of \textit{neural machine translation} (NMT), which is paradigmatically close to the tasks typically considered for compositionality tests, where the target represents the `meaning' of the input.\footnote{E.g. SCAN's inputs are instructions (\exa{walk twice}) with executions as outputs (\exa{walk walk}) \citep{lake2018generalization}.}
Furthermore, MT is an important domain of NLP, for which compositional generalisation is important to produce more robust translations and train adequate models for low-resource languages \citep[see, e.g.][]{chaabouni2021can}.
As an added advantage, compositionality is traditionally well studied and motivated for MT \citep{rosetta1994rosetta,janssen1997compositionality,janssen1998algebraic}.

We reformulate three theoretically grounded tests from \citet{hupkes2020compositionality}: \emph{systematicity}, \emph{substitutivity} and \emph{overgeneralisation}.
Since accuracy -- commonly used in artificial compositionality tests -- is not a suitable evaluation metric for MT, we base our evaluations on the extent to which models behave \emph{consistently}, rather than correctly.
In our tests for systematicity and substitutivity, we consider whether processing is \emph{local}; in our overgeneralisation test, we consider how models treat idioms that are assumed to require \emph{global} processing.

Our results indicate that models often do not behave compositionally under the local interpretation, but exhibit behaviour that is \emph{too local} in other cases.
In other words, models have the ability to process phrases both locally and globally but do not always correctly modulate between them.
We further show that some inconsistencies reflect variation in natural language, whereas others are true \emph{compositional mistakes}, exemplifying the need for both local and global compositionality as well as illustrating the need for tests that encompass them both.

With our study, we contribute to ongoing questions about the compositional abilities of neural networks, and we provide nuance to the nature of this question when natural language is concerned: how local should the compositionality of models for natural language actually be?
Aside from an empirical study, our work is also a call to action: we should rethink the evaluation of compositionality in neural networks and develop benchmarks using \emph{real} data to evaluate compositionality on natural language, where composing meaning is not as straightforward as doing the math.

%% file: tables/template_table.tex
\begin{table*}[!h]
\small
\centering
\resizebox{1.94\columnwidth}{!}{
\begin{subtable}[b]{\columnwidth}
\begin{tabular}{cll}
\toprule
\vspace{-0.5mm}
$n$ & \textbf{Template} \\ \midrule
1   & The \bl{N}\SPSB{}{people} \bl{V} the \bl{N}\SPSB{sl}{people} . \\
2   & The \bl{N}\SPSB{}{people} \bl{Adv} \bl{V} the \bl{N}\SPSB{sl}{people} . \\ 
3   & The \bl{N}\SPSB{}{people} \bl{P} the \bl{N}\SPSB{sl}{vehicle} \bl{V} the \bl{N}\SPSB{sl}{people} . \\ 
4   & The \bl{N}\SPSB{}{people} and the \bl{N}\SPSB{}{people} \bl{V} the \bl{N}\SPSB{sl}{people} . \\
5   & The \bl{N}\SPSB{sl}{quantity} of \bl{N}\SPSB{pl}{people} \bl{P} the \bl{N}\SPSB{sl}{vehicle} \bl{V} the \bl{N}\SPSB{sl}{people} . \\
6   & The \bl{N}\SPSB{}{people} \bl{V} that the \bl{N}\SPSB{pl}{people} \bl{V}. \\
7   & The \bl{N}\SPSB{}{people} \bl{Adv} \bl{V} that the \bl{N}\SPSB{pl}{people} \bl{V} . \\
8   & The \bl{N}\SPSB{}{people} \bl{V} that the \bl{N}\SPSB{pl}{people} \bl{V} \bl{Adv} . \\
9   & The \bl{N}\SPSB{}{people} that \bl{V} \bl{V} the \bl{N}\SPSB{sl}{people} . \\
10  & The \bl{N}\SPSB{}{people} that \bl{V} \bl{Pro} \bl{V} the \bl{N}\SPSB{sl}{people} . \\
    \bottomrule
    \end{tabular}
    \caption{Synthetic templates\vspace{-1.5mm}}
    \label{tab:synthetic_data}
\end{subtable}\begin{subtable}[b]{\columnwidth}
\small
\centering
\begin{tabular}{cll}
\toprule
\vspace{-0.5mm}
$n$ & \textbf{Template} \\ \midrule
1,2,3 & The \bl{N}\SPSB{}{people} \textcolor{red}{VP}\SBR{1,2,3} . \\
      & \textit{The men are gon na have to move off-camera .} \\
4,5   & The \bl{N}\SPSB{}{people} read(s) an article about \textcolor{red}{NP}\SBR{1,2} . \\ 
      & \textit{The man reads an article about the development} \\
      & \textit{of ascites in rats with liver cirrhosis .} \\
6,7   & An article about \textcolor{red}{NP}\SBR{3,4} is read by \bl{N}\SPSB{}{people} . \\ 
      & \textit{An article about the criterion on price stability ,} \\
      & \textit{which was 27 \% , is read by the child .} \\
8,9,10& Did the \bl{N}\SPSB{}{people} hear about \textcolor{red}{NP}\SBR{5,6,7} ? \\
      & \textit{Did the teacher hear about the march on} \\
      & \textit{Employment which happened here on Sunday ?} \\
    \bottomrule
    \end{tabular}
    \caption{Semi-natural templates\vspace{-1.5mm}}
    \label{tab:semi_natural}
\end{subtable}
}
\caption{The synthetic and semi-natural templates, with POS tags of the lexical items varied shown in blue with the plurality as superscript and the subcategory as subscript.  The OPUS-extracted NP and VP fragments are red.}
\vspace{-0.3cm}
\end{table*}

%% file: background.tex
\section{Local and global compositionality}
\label{sec:LoC}

Tests for compositional generalisation in neural networks typically assume an arithmetic-like version of compositionality, in which meaning can be computed bottom up.
The compositions require only local information -- they are context independent and unambiguous:  \exa{walk twice after jump thrice} \citep[a fragment from SCAN by][]{lake2018generalization} is evaluated similarly to $(2 + 1) \times (4 - 5)$.
In MT, this type of compositionality would imply that a change in a word or phrase should affect only the translation of that word or phrase, or at most the smallest constituent it is a part of.
For instance, the translation of \exa{the girl} should not change depending on the verb phrase that follows it, and in the translation of a conjunction of two sentences, making a change in the first conjunct should not change the translation of the second.
While translating in such a local way seems robust and productive, it is not always realistic -- e.g.\ consider the translation of 
\exa{dates} in \exa{She hated bananas and she liked dates}.\looseness=-1

In linguistics and philosophy of language, the \textit{level} of compositionality has been widely discussed, which led to a variety of definitions.
One of the most well-known ones is from \citet[][]{partee1984compositionality}:

\vspace{-.5mm}
\begin{quote}
``The meaning of a compound expression is a function of the meanings of its parts and of the way they are syntactically combined.''\footnote{This straightforwardly extends to translation, by replacing \emph{meaning} with \emph{translation} \citep{rosetta1994rosetta}.}
\end{quote}

\vspace{-.5mm}
\noindent This definition hardly places restrictions on the relationship between expressions and their parts.
The type of function that relates them is unspecified and could take into account the global syntactic structure or external arguments, and the parts' meanings can depend on global information.
\citeauthor{partee1984compositionality}'s definition is therefore called \textit{weak}, \textit{global}, or \textit{open} compositionality \citep{szabo2012case, garcia2019open}.
When, instead, the meaning of a compound depends only on the meanings of its largest parts, regardless of their internal structure (similar to arithmetic),
that is \emph{strong}, \emph{local} or \textit{closed} compositionality \citep{jacobson2002dis, szabo2012case}.
Under the local interpretation, natural language can hardly be considered compositional -- many frequent phenomena such as homonyms, idioms and scope ambiguities cannot be resolved locally \citep{pagin2010compositionality, pavlick2016most}.
The global interpretation handles such cases straightforwardly but does not match up with many a person's intuitions about the compositionality of language.
After all, how useful is compositionality if composing the meanings of parts requires the entire rest of the sentence?
This paradox inspired debates on the compositionality of natural language and is also highly relevant in the context of evaluating compositionality in neural models.

Previous compositionality tests (\S\ref{subsec:related_work}) considered only the local interpretation of compositionality, but to what extent is that relevant given the type of compositionality actually required to model natural language?
Here, we aim to open up the discussion about what it means for computational models of language to be compositional by considering properties that require composing meaning locally as well as globally and evaluating them in models trained on unadapted natural language corpora.

%% file: setup.tex
\section{Setup}

\input{model}

\input{data}

%% file: model.tex
\subsection{Model and training}\label{sec:model}

We focus on English-Dutch translation, for which we can ensure good command for both languages.
We train Transformer-base models \citep{vaswani2017attention} using Fairseq \citep{ott2019fairseq}.
Our training data consists of a collection of MT corpora bundled in \textsc{OPUS} \citep{tiedemann2020opus}, of which we use the English-Dutch subset provided by \citet{tiedemann-2020-tatoeba}, which contains 69M sentence pairs.\footnote{Visit \href{https://github.com/Helsinki-NLP/Tatoeba-Challenge/blob/master/data/README-v2020-07-28.md}{the Tatoeba challenge} for the \textsc{OPUS} training data.}
To examine the impact of the amount of training data -- a dimension that is relevant because compositionality is hypothesised to be more important when resources are scarcer -- we train one setup using the \textbf{full} dataset, one using $\frac{1}{8}$ of the data (\textbf{medium}), and one using one million source-target pairs in the \textbf{small} setup. 
For each setup, we train models with five seeds and average the results.

\begin{figure*}[!h]
    \centering
    \begin{subfigure}[b]{0.45\columnwidth}
        \includegraphics[width=\columnwidth]{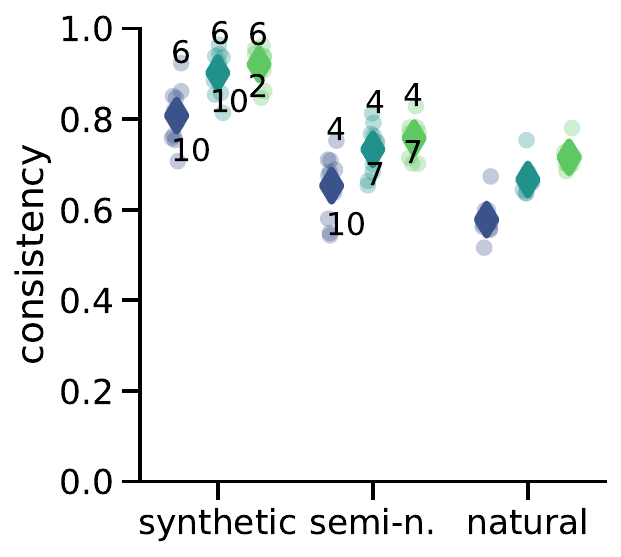}
        \caption{$\text{S}_1 \rightarrow \text{S}_1^\prime$}
        \label{fig:systematicity_sconj_s1}
    \end{subfigure}
    \begin{subfigure}[b]{0.45\columnwidth}
        \includegraphics[width=\columnwidth]{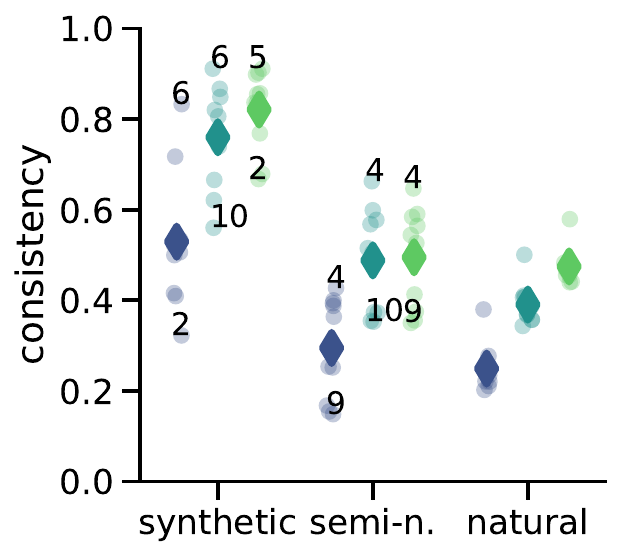}
        \caption{$\text{S}_1 \rightarrow \text{S}_3$}
        \label{fig:systematicity_sconj_s3}
    \end{subfigure}
    \begin{subfigure}[b]{0.45\columnwidth}
        \includegraphics[width=\columnwidth]{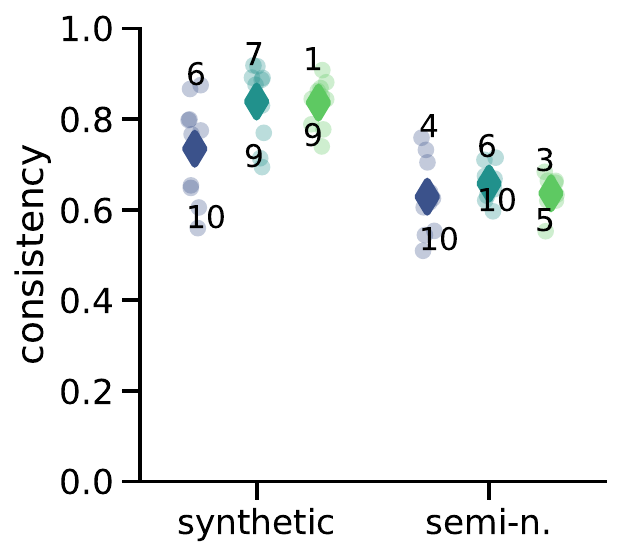}
        \caption{$\text{NP} \rightarrow \text{NP}^\prime$}
        \label{fig:systematicity_snpvp_np}
    \end{subfigure}
    \begin{subfigure}[b]{0.49\columnwidth}
        \includegraphics[width=\columnwidth]{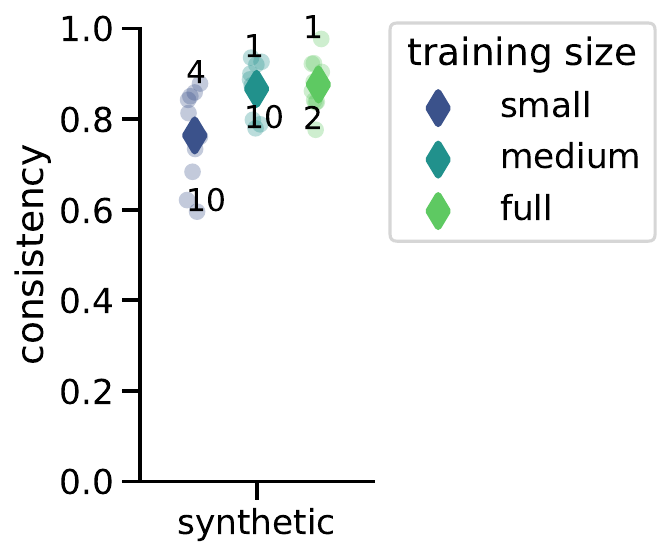}
        \caption{$\text{VP} \rightarrow \text{VP}^\prime$\hspace{8mm}}
        \label{fig:systematicity_snpvp_vp}
    \end{subfigure}
    \caption{Systematicity results for setup \texttt{\small S\;$\rightarrow$\;S\;CONJ\;S} (a and b) and \texttt{\small S\;$\rightarrow$\;NP\;VP} (c and d).
    Consistency scores are shown per evaluation data type ($x$-axis) and training dataset size (colours).
    Data points represent templates ($\circ$) and means over templates ($\diamond$).}
    \label{fig:systematicity}
\vspace{-0.3cm}
\end{figure*}

To evaluate our trained models, we adopt \textsc{Flores-101} \citep{goyal2021flores}, which contains 3001 sentences from Wikinews, Wikijunior and WikiVoyage, translated by professional translators, split across three subsets.
We train the models until convergence on the `dev' set.
Afterwards, we compute SacreBLEU scores on the `devtest' set \citep{post2018call}, using beam search (beam size = 5), yielding scores of $20.6\!\pm\!.4$, $24.4\!\pm\!.3$ and $25.8\!\pm\!.1$ for the small, medium and full datasets, respectively.\footnote{All training details are listed in Appendix~\ref{app:reproducibility}.}

\begin{figure}[t]\small
\includegraphics[width=\columnwidth]{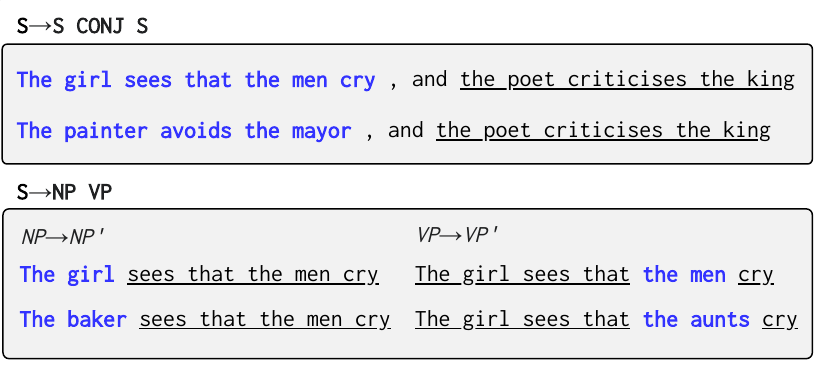}
\caption{Illustration of the systematicity experiments \texttt{S\;$\rightarrow$\;S\;CONJ\;S} ($\text{S}_1$\;$\rightarrow$\;$\text{S}_3$ is shown) and \texttt{S\;$\rightarrow$\;NP\;VP} (both versions are shown).
Each experiment involves extracting translations before and after the replacement of the blue part, and then comparing the translation of the underlined words.}
\label{fig:systematicity_explanation}
\vspace{-0.3cm}
\end{figure}

%% file: data.tex
\subsection{Evaluation data}\label{sec:data}

While all our models are trained on fully natural data, for evaluation we use different types of data: synthetic, semi-natural and natural data.

\paragraph{Synthetic data}
For our \textbf{synthetic} evaluation data, we consider the data generated by \citet{lakretz2019emergence},
previously used to probe for hierarchical structure in neural language models.
This data consist of sentences with a fixed syntactic structure and diverse lexical material.
We extend the vocabulary and the templates used to generate the data and generate 3000 sentences for each of the resulting 10 templates (see Table~\ref{tab:synthetic_data}).

\paragraph{Semi-natural data}
In the synthetic data, we have full control over the sentence structure and lexical items, but the sentences are shorter (9 tokens vs 16 in \textsc{OPUS}) and simpler than typical in NMT data.
To obtain more complex yet plausible test sentences, we employ a data-driven approach to generate \textbf{semi-natural} data. 
Using the tree substitution grammar Double DOP \citep{vancranenburgh2016disc}, we obtain noun and verb phrases (NP, VP) whose structures frequently occur in \textsc{OPUS}.
We then embed these NPs and VPs in ten synthetic templates with 3000 samples each (see Table~\ref{tab:semi_natural}).
See Appendix~\ref{ap:ddop} for details on the data generation.

\paragraph{Natural data}
Lastly, we extract \textbf{natural} data directly from \textsc{OPUS}, as detailed in the subsections of the individual tests (\S\ref{sec:experiments}).

%% file: tests.tex
\section{Experiments and results}
\label{sec:experiments}

In our experiments, we consider \emph{systematicity} (\S\ref{subsec:systematicity}) and \emph{substitutivity} (\S\ref{subsec:substitutivity}), to test for local compositionality, 
and \emph{idiom translation} 
to probe for a more global type of processing (\S\ref{subsec:global_compositionality}).

\begin{figure*}[!ht]
    \centering
    \begin{subfigure}[b]{\columnwidth}
    \includegraphics[width=\columnwidth]{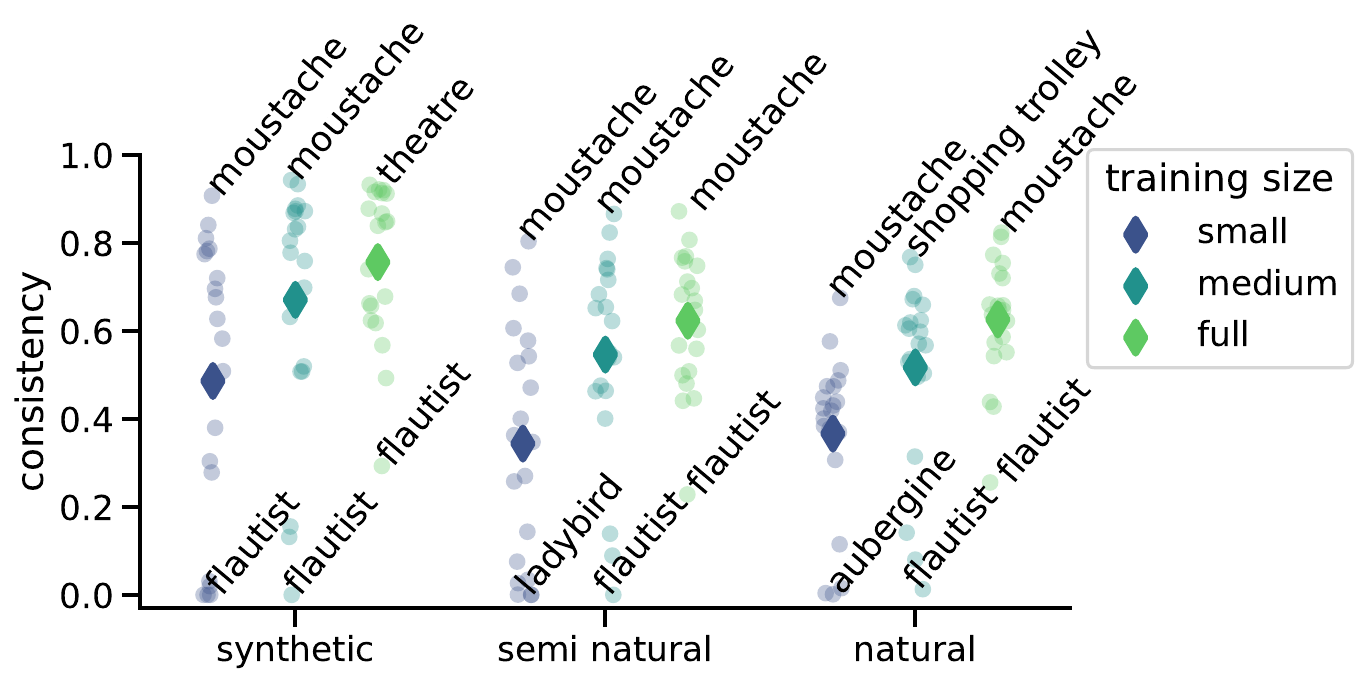}
    \caption{}
        \vspace{-2mm}
    \label{fig:substitutivity}
    \end{subfigure}
    \begin{subfigure}[b]{\columnwidth}
\includegraphics[width=\columnwidth]{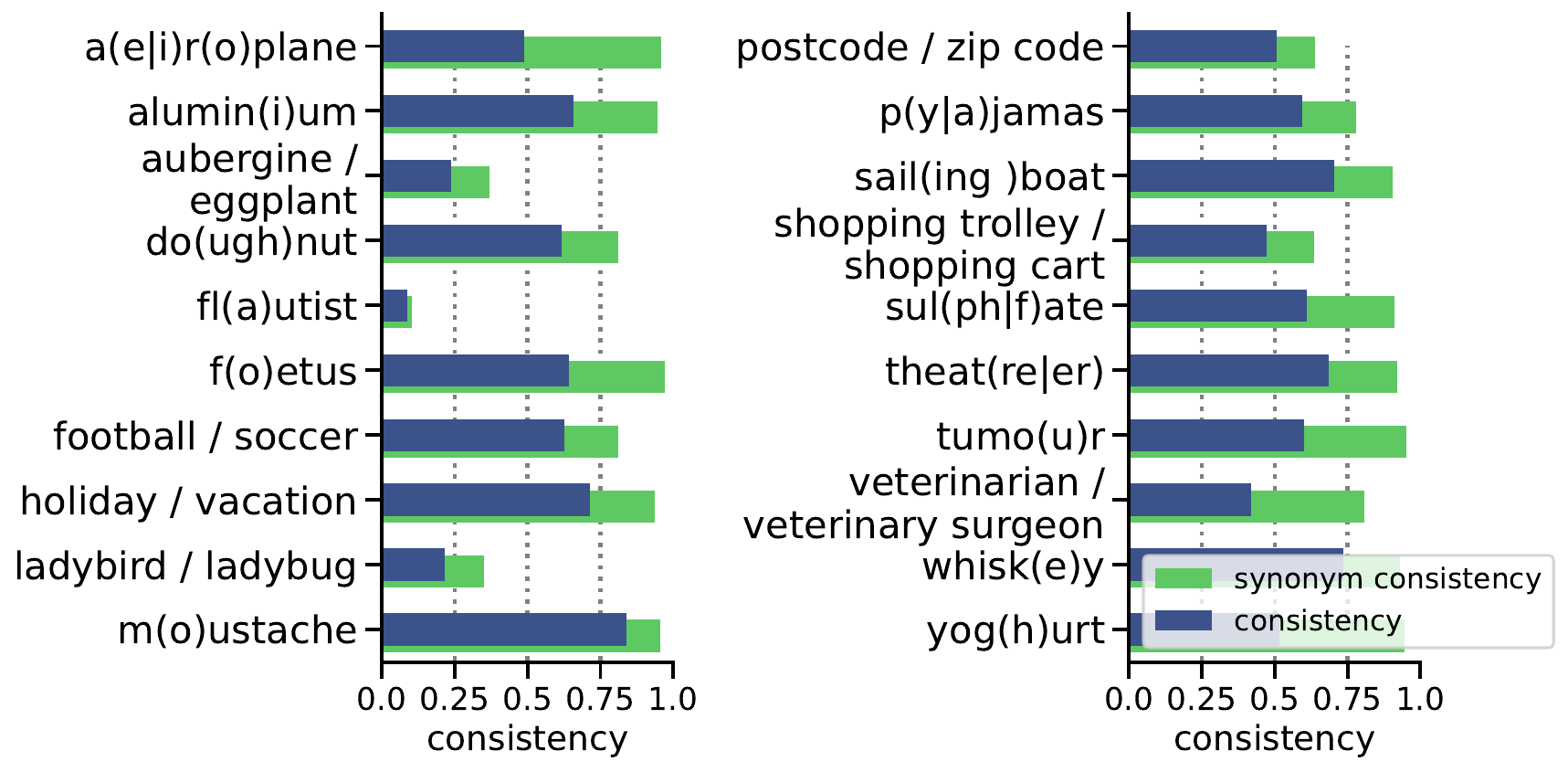}
        \caption{}
        \vspace{-2mm}
\label{fig:per_synonym}
    \end{subfigure}
    \caption{(a) Consistency scores of synonyms (averaged $\diamond$, and per synonym $\circ$) for substitutivity per evaluation data type, for three training set sizes.
    (b) Consistency per synonym, measured using full sentences (in dark blue) or the synonym's translation only (in green), averaged over training dataset sizes and data types.}
\vspace{-0.3cm}
\end{figure*}

\input{systematicity}

\input{substitutivity}

\input{global_compositionality}

%% file: systematicity.tex
\subsection{Systematicity}
\label{subsec:systematicity}

One of the most commonly tested properties of compositional generalisation is \textbf{systematicity} -- the ability to understand novel combinations made up from known components \citep[most famously,][]{lake2018generalization}.
In natural data, the number of potential recombinations to consider is infinite.
We chose to focus on recombinations in two sentence-level context-free rules: \texttt{S\;$\rightarrow$\;NP\;VP} and \texttt{S\;$\rightarrow$\;S\;CONJ\;S}.

\subsubsection{Experiments}
\paragraph{Test design}
\label{subsec:systematicity_test_design}
The first setup, \texttt{S\;$\rightarrow$\;NP\;VP}, concerns recombinations of noun and verb phrases.
We extract translations for input sentences from the templates from \S\ref{sec:data}, as well as versions of them with the (1) noun (NP $\rightarrow$ NP') or (2) verb phrase (VP $\rightarrow$ VP') adapted.
In (1), a noun from the NP in the subject position is replaced with a different noun while preserving number agreement with the VP.
In (2), a noun in the VP is replaced.
NP $\rightarrow$ NP' is applied to both synthetic and semi-natural data; VP $\rightarrow$ VP' only to synthetic data.
We use 500 samples per template per condition per data type.

The second setup, \texttt{S\;$\rightarrow$\;S\;CONJ\;S}, involves phrases concatenated using \exa{and}, and tests whether the translation of the second sentence is dependent on the first sentence.
We concatenate two sentences ($\text{S}_1$ and $\text{S}_2$) from different templates, and we consider again two different conditions.
First, in condition $\text{S}_1\rightarrow\text{S}^\prime_1$, we make a minimal change to $\text{S}_1$ yielding $\text{S}^\prime_1$ by changing the noun in its verb phrase.
In $\text{S}_1\rightarrow\text{S}_3$, instead, we replace $\text{S}_1$ with a sentence $\text{S}_3$ that is sampled from a template different from $\text{S}_1$.
We compare the translation of $\text{S}_2$ in all conditions.
For consistency, the first conjunct is always sampled from the synthetic data templates. 
The second conjunct is sampled from synthetic data, semi-natural data, or from natural sentences sampled from \textsc{OPUS} with similar lengths and word-frequencies as the semi-natural inputs.
We use 500 samples per template per condition per data type.
Figure~\ref{fig:systematicity_explanation} provides an illustration of the different setups experimented with.

\paragraph{Evaluation}
In artificial domains, systematicity is evaluated by leaving out combinations of `known components' from the training data and using them for testing purposes.
The necessary familiarity of the components (the fact that they are `known') is ensured by high training accuracies, and systematicity is quantified by measuring the test set accuracy.
If the training data is a natural corpus and the model is evaluated with a measure like BLEU in MT, this strategy is not available.
We observe that being systematic requires being consistent in the interpretation assigned to a (sub)expression across contexts, both in artificial and natural domains.
Here, we, therefore, focus on \textbf{consistency} rather than accuracy, allowing us to employ a model-driven approach that evaluates the model's systematicity as the consistency of the translations when presenting words or phrases in multiple contexts.

We measure consistency as the equality of two translations after accounting for anticipated changes.
For instance, in the \texttt{S\;$\rightarrow$\;NP\;VP} setup, two translations are consistent if they differ in one word only, after accounting for determiner changes in Dutch (\exa{de} vs\ \exa{het}).
In the evaluation of \texttt{S\;$\rightarrow$\;S\;CONJ\;S}, we measure the consistency of the translations of the second conjunct.

\subsubsection{Results}
Figure~\ref{fig:systematicity} shows the results for the \texttt{S\;$\rightarrow$\;NP\;VP} and \texttt{S\;$\rightarrow$\;S\;CONJ\;S} setups (numbers available in Appendix~\ref{ap:systematicity}).
The average performance for the natural data closely resembles the performance on \textit{semi-}natural data, suggesting that the increased degree of control did not severely impact the results obtained using this generated data.\footnote{In our manual analysis (\S\ref{sec:manual_analysis}), however, we did observe a slightly different distribution of changes between these setups.}
In general, the consistency scores are low, illustrating that models are prone to changing their translation of a (sub)sentence after small (unrelated) adaptations to the input.
It hardly matters whether that change occurs in the sentence itself (\texttt{S\;$\rightarrow$\;NP\;VP}), or in the other conjunct (\texttt{S\;$\rightarrow$} \texttt{S\;CONJ\;S}), suggesting that the processing of the models is not local as assumed in strong compositionality.
Models trained on more data seem more locally compositional, a somewhat contradictory solution to achieving compositionality, which, after all, is assumed to underlie the ability to generalise usage from \emph{few} examples \citep{lake2019human}.
This trend is also at odds with the hypothesis that inconsistencies are a consequence of the natural variation of language, which models trained on \emph{more} data are expected to better capture.

%% file: substitutivity.tex
\subsection{Substitutivity}
\label{subsec:substitutivity}

Under a local interpretation of the principle of compositionality, synonym substitutions should be meaning-preserving: substituting a constituent in a complex expression with a synonym should not alter the complex expression's meaning, or, in the case of MT, its translation.
Here, we test to what extent models' translations abide by this principle, by performing the \textbf{substitutivity} test from \citet{hupkes2020compositionality}, that measures whether the outputs remain consistent after synonym substitution.

\subsubsection{Experiments}
To find synonyms -- source terms that translate into the same target terms -- we exploit the fact that OPUS contains texts both in British and American English.
Therefore, it contains synonymous terms that are spelt different -- e.g.\ \exa{doughnut} / \exa{donut} -- and synonymous terms with a very different form -- e.g.\ \exa{aubergine} / \exa{eggplant}.
We use 20 synonym pairs in total (see Figure~\ref{fig:per_synonym}).

\paragraph{Test design}
Per synonym pair, we select natural data from OPUS in which the terms appear and perform synonym substitutions.
Thus, each sample has two sentences, one with the British and one with the American English term.
We also insert the synonyms into the synthetic and semi-natural data using 500 samples per synonym pair per template, through subordinate clauses that modify a noun -- e.g. ``the king \textit{that eats the doughnut}''.
In Appendix~\ref{ap:substitutivity}, Table~\ref{tab:substitutivity_appendix}, we list all clauses used.

\paragraph{Evaluation}
Like systematicity, we evaluate substitutivity using the consistency score, expressing whether the model translations for a sample are identical.
We report both the full sentence consistency and the consistency of the synonyms' translations only, excluding the context.
Cases in which the model omits the synonym from both translations are labelled as consistent if the rest of the translation is the same for both input sequences.

\subsubsection{Results}
In Figure~\ref{fig:substitutivity}, we summarise all substitutivity consistency scores (tables are in Appendix~\ref{ap:substitutivity}).
We observe trends similar to the systematicity results: models trained on larger training sets perform better and synthetic data yields more consistent translations compared to (semi-)natural data.
We further observe large variations across synonyms, for which we further detail the performance aggregated across experimental setups in Figure~\ref{fig:per_synonym}.
The three lowest scoring synonyms -- \exa{flautist}, \exa{aubergine} and \exa{ladybug} -- are among the least frequent synonyms (see Appendix~\ref{ap:substitutivity}), which stresses the importance of frequency for the model to pick up on synonymy.

In Figure~\ref{fig:per_synonym}, we show both the regular consistency and the consistency of the synonym translations, illustrating that a substantial part of the inconsistencies are due to varying translations of the context rather than the synonym itself, stressing again the non-local processing of the models.

%% file: global_compositionality.tex
\subsection{Global compositionality}
\label{subsec:global_compositionality}

In our final test, we focus on exceptions to compositional rules.
In natural language, typical exceptions that constitute a challenge for local compositionality are \emph{idioms}.
For instance, the idiom ``raining cats and dogs'' should be treated globally to arrive at its meaning of heavy rainfall.
A local approach would yield an overly literal, non-sensical translation (``het regent katten en honden'').
When a model's translation is too local, we follow \citet{hupkes2020compositionality} in saying that it \textbf{overgeneralises}, or, in other words, it applies a general rule to an expression that is an exception to this rule.
Overgeneralisation indicates that a language learner has internalised the general rule \citep[e.g.][]{penke2012dual}.

\subsubsection{Experiments}
We select 20 English idioms for which an accurate Dutch translation differs from the literal translation from the English MAGPIE corpus \citep{haagsma2020magpie}.
Because acquisition of idioms is dependent on their frequency in the corpus, we use idioms with at least 200 occurrences in OPUS based on exact matches, for which over 80\% of the target translations does not contain a literal translation.

\paragraph{Test design}
Per idiom, we extract \textit{natural} sentences containing the idiom from OPUS. 
For the synthetic and semi-natural data types, we insert the idiom in 500 samples per idiom per template, by attaching a subordinate clause to a noun -- e.g.\ ``the king \emph{that said `I knew the formula \textbf{by heart}'}''. 
The clauses used can be found in Appendix~\ref{ap:global_compositionality}, Table~\ref{tab:overgeneralisation_appendix}.

\paragraph{Evaluation}
Per idiom, we assess how often a model overgeneralises and how often it translates the idiom globally. 
To do so, we identify keywords that indicate that a translation is translated locally (literal) instead of globally (idiomatic).
If the keywords' literal translations are present, the translation is labelled as an overgeneralised translation. 
For instance, for ``by heart'', the presence of ``hart'' (``heart'') suggests a literal translation. An adequate paraphrase would say ``uit het hoofd'' (``from the head'').
See Appendix~\ref{ap:global_compositionality}, Table~\ref{tab:overgeneralisation_appendix}, for the full list of keywords.
We evaluate overgeneralisation for ten intermediate training checkpoints.

\begin{figure}
    \centering
    \begin{subfigure}[b]{\columnwidth}\centering
    \includegraphics[width=\textwidth]{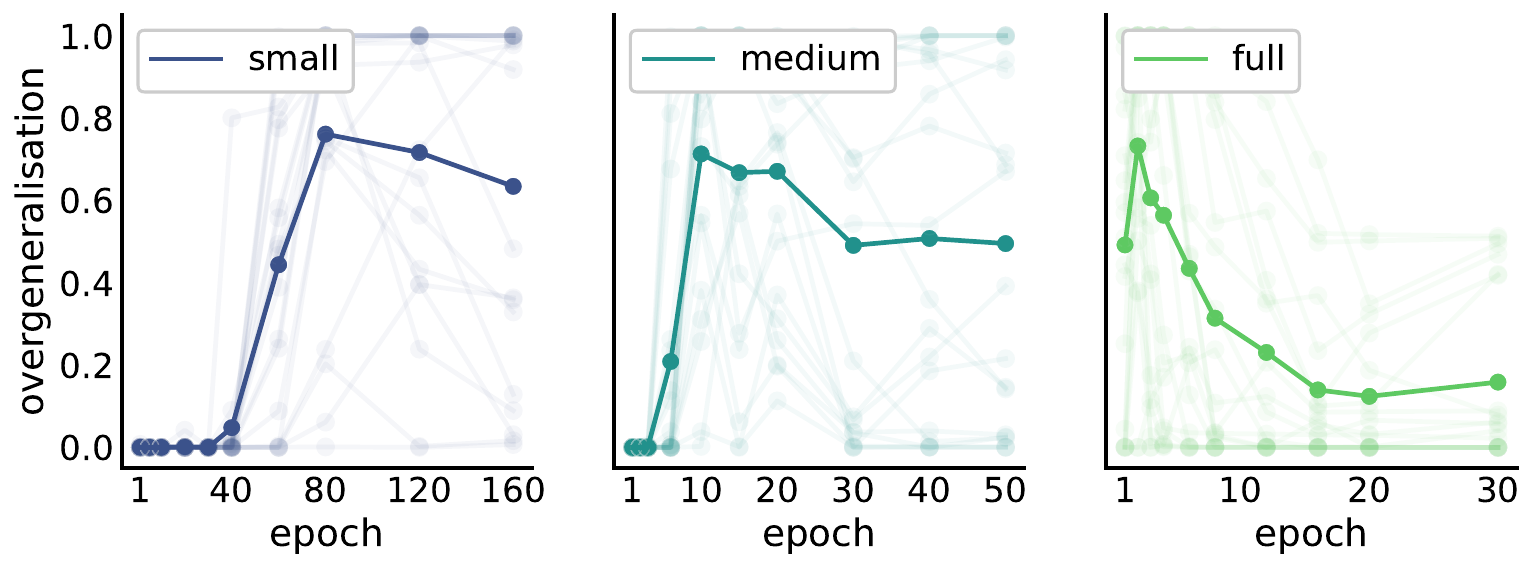}
    \caption{Synthetic}
    \end{subfigure}
    \begin{subfigure}[b]{\columnwidth}\centering
    \includegraphics[width=\textwidth]{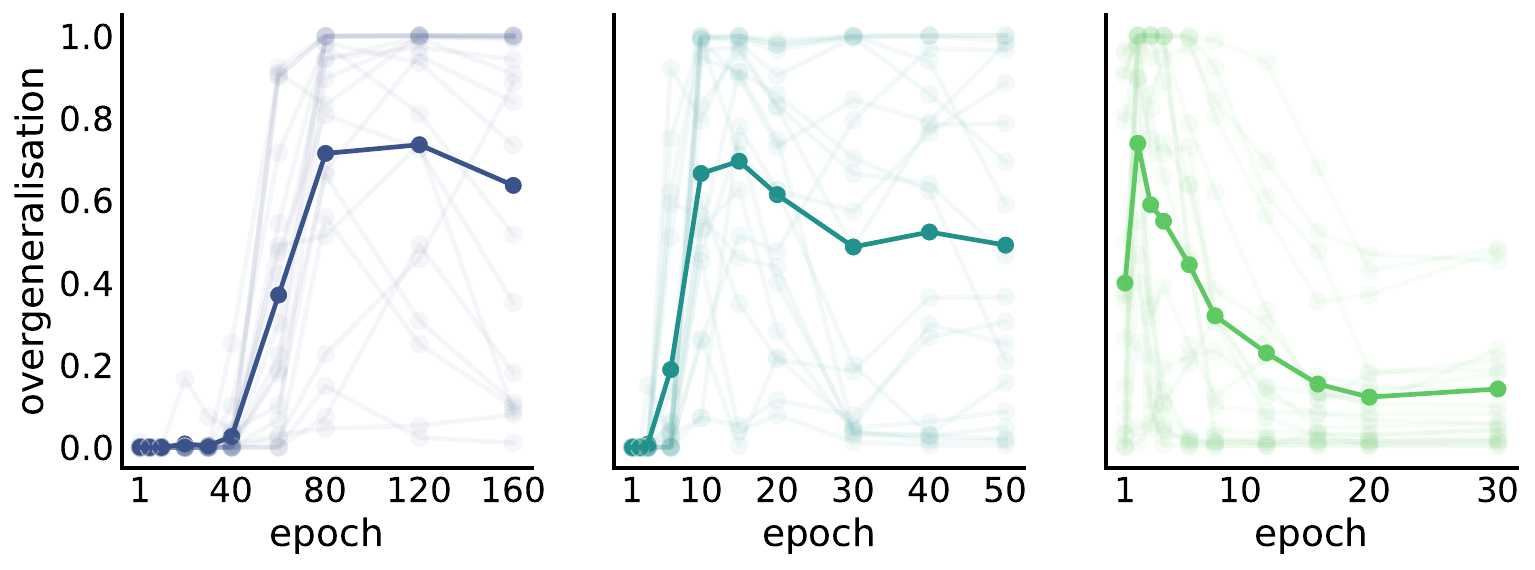}
    \caption{Semi-Natural}
    \end{subfigure}
    \begin{subfigure}[b]{\columnwidth}\centering
    \includegraphics[width=\textwidth]{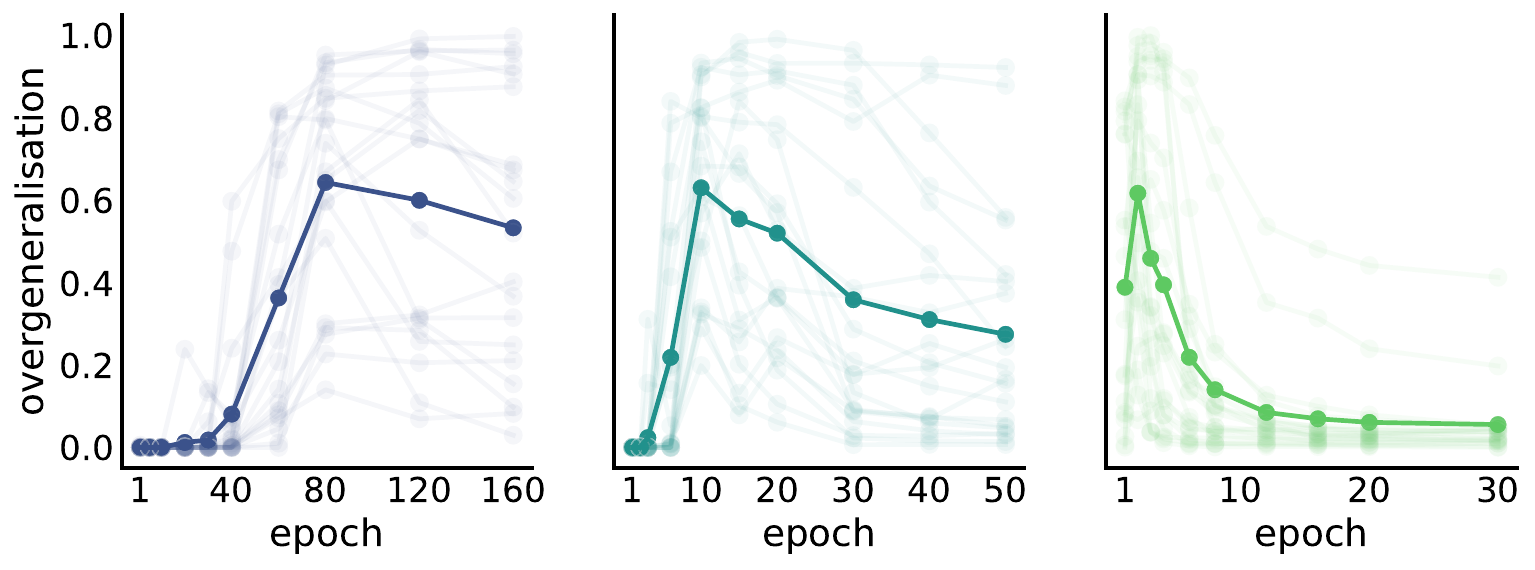}
    \caption{Natural}
    \end{subfigure}
    \caption{Visualisation of overgeneralisation for idioms throughout training, with a line per idiom and the overall mean. Overgeneralisation occurs early on in training and precedes memorisation of idioms' translations.
    The colours indicate different training dataset sizes.}
    \label{fig:global_compositionality}
    \vspace{-0.3cm}
\end{figure}

\subsubsection{Results}
In Figure~\ref{fig:global_compositionality}, we report our results.\footnote{Note that epochs consist of different numbers of samples: 1M, 8.6M and 69M for small, medium and full. Appendix~\ref{ap:global_compositionality} further details numerical results per idiom.}
For all evaluation data types and all training set sizes, three phases can be identified.
Initially, the translations do not contain the idiom's keyword, not because the idiom's meaning is paraphrased in the translation, but because the translations consist of high-frequency words in the target language only. 
Afterwards, overgeneralisation peaks: the model emits a very literal translation of the idiom.
Finally, the model starts to memorise the idiom's translation.
This is in accordance with results from \citet{hupkes2020compositionality}, and earlier results presented in the past tense debate by -- among others -- \citet{rumelhart1986learning}.

Although the height of the overgeneralisation peak is similar across evaluation data types and training set sizes, overgeneralisation is more prominent in converged models trained on smaller datasets than it is in models trained on the full corpus.\footnote{Convergence is based on BLEU scores for validation data.}
In addition to training dataset size, the type of evaluation data used also matters: there is more overgeneralisation for synthetic and semi-natural data compared to natural data, stressing the impact of the context in which an idiom is embedded.
The extreme case of a context unsupportive of an idiomatic interpretation is a sequence of random words. To evaluate the hypothesis that this yields local translations, we surround the idioms with ten random words.
The results (Appendix~\ref{ap:global_compositionality}, Table~\ref{tab:overgeneralisation_appendix}) indicate that, indeed,  when the context provides no support at all for a global interpretation, the model provides a local translation for nearly all idioms.
Overall, the results of this test provide an interesting contrast with our substitutivity and systematicity results: where in those tests, we saw processing that was \emph{less local} than we expected, here, the behaviour shown by the models is instead \emph{not global enough}.

%% file: manual_analysis.tex
\section{Manual analysis}
\label{sec:manual_analysis}

Our systematicity and substitutivity results demonstrate that models are not behaving compositional according to a strict definition of compositionality.
However, we ourselves have argued that strict compositionality is not always appropriate to handle natural language.
A reasonable question to ask is thus: are the inconsistencies we marked as non-compositional actually incorrect?

\begin{figure}
	\begin{subfigure}[b]{\columnwidth}
		\includegraphics[width=\textwidth]{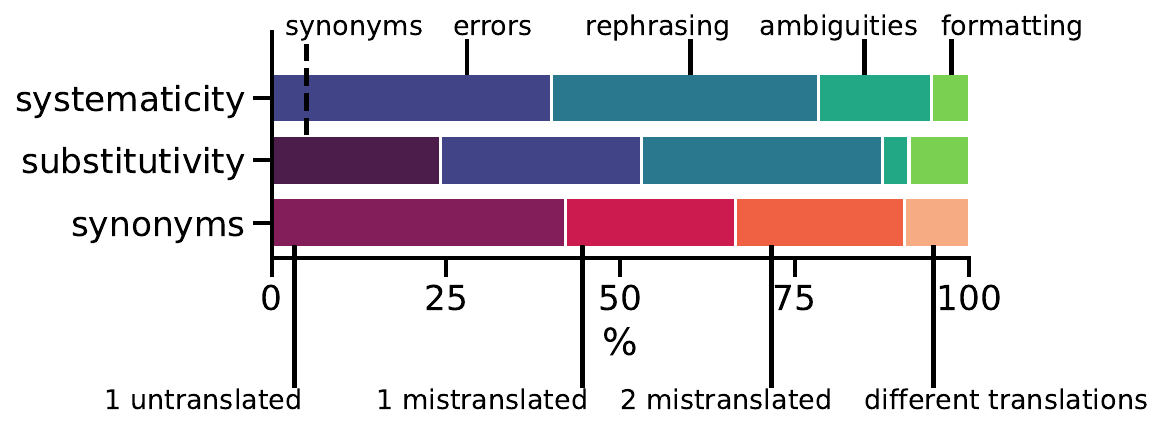}
	\end{subfigure}
	\caption{Relative frequencies of manually labelled inconsistencies in translations, averaged over data types and training set sizes.
	         The `synonyms' distribution further details the category `synonyms' from row two.}
	\label{fig:manual_analysis_summary}
	\vspace{-.3cm}
\end{figure}

\paragraph{Annotation setup} 
To address this question, we perform a manual analysis.
We annotate 900 inconsistent translation pairs of the systematicity and substitutivity tests to establish whether the inconsistencies are benign or concerning.
We consider four different types of changes:
\begin{enumerate}[noitemsep,topsep=0pt]
\item cases of \textit{rephrasing}, where both translations are equally (in)correct;
\item changes reflecting different interpretations of \textit{source ambiguities};
\item cases in which one of the two translations contains an \textit{error};
\item \textit{formatting} (mostly punctuation) changes.
\end{enumerate}
For substitutivity samples, we also annotate whether the changes are related to the translation of the synonym, where we distinguish cases where
\begin{enumerate}[noitemsep,topsep=0pt]
\item[i.] one of the synonym translations is incorrect;
\item[ii.] both are incorrect but in a different manner;
\item[iii.] both are correct but translated differently;
\item[iv.] one synonym remains untranslated.
\end{enumerate}
We annotate all changes observed per pair and report the relative frequency per class. 
We summarise the results, aggregated over different training set sizes and the three data types, in Figure~\ref{fig:manual_analysis_summary}.
For a more elaborate analysis and a breakdown per model and data type, we refer to Appendix~\ref{app:man_analysis}.

\paragraph{Results}
In the systematicity test, 40\% of the marked inconsistencies reflects wrongfully translated parts in one of the two sentences, whereas 38\% contains examples of rephrasing, 16\% reflects ambiguities in the source sentences and 6\% is caused by formatting differences. 
For substitutivity, most inconsistencies are similar to the ones observed in systematicity: only 24\% involves the synonyms' translations, where one of them being untranslated was the most frequent category.
 
The distribution of these types of inconsistencies differ strongly per training data type.
For models trained on less data, inconsistencies are more likely to represent errors, whereas models trained on more data rephrase more often. 
This result emphasises that for lower-resource settings, being compositional is particularly relevant.

Another demonstration of this relevance comes from the observation that although models \textit{can} emit correct translations for nearly all synonyms,\footnote{Apart from the model with the small training dataset that cannot translate \exa{flautist} and \exa{ladybug}.}
they do not always do so, depending on the context.
To give a peculiar example: in ``The child admires the king that eats the \{doughnut, donut\}'', the snack was occasionally translated as ``ezel'' (\exa{donkey}).

\paragraph{Robustness and predictability}
Finally, we would like to stress that while rephrasing often might seem benign rather than concerning from the perspective of emitting adequate translations, its harmlessness still deserves some thought.
There is a fine line between rephrasing and mistranslating: whether ``the \textit{single largest} business establishment'' is referred to as ``de grootste'' (\exa{the largest}) or ``de enige grootste'' (\exa{the only largest}) may make or break a translation. 
Furthermore, if changes are unrelated to the contextual change (e.g. replacing ``soccer'' with ``football''), this can be undesirable from a robustness and reliability perspective.
This point becomes even more pronounced in cases where both translations are correct but have a different meaning.
To analyse the extent to which inconsistencies are actually unmotivated, we investigated if we could trace them back to the contextual change, in particular focusing on whether changing synonyms from British to American spelling or vice versa might trigger a change in style or tone.
We could not find evidence of such motivations, indicating that even correct cases of rephrasing were not caused by contextual changes that were \emph{necessary} to take into account.

%% file: related_work.tex
\section{Related work}
\label{subsec:related_work}

In previous work, a variety of artificial tasks have been proposed to evaluate compositional generalisation using non-i.i.d.\ test sets that are designed to assess a specific characteristic of compositional behaviour.
Examples are \emph{systematicity} \citep{lake2018generalization, bastings2018jump, hupkes2020compositionality}, \emph{substitutivity} \citep{mul2019siamese,hupkes2020compositionality}, \emph{localism} \citep{hupkes2020compositionality,saphra-lopez-2020-lstms}, \emph{productivity} \citep{lake2018generalization} or \emph{overgeneralisation} \citep{korrel2019transcoding,hupkes2020compositionality,dankers-etal-2021-generalising}. 
Generally, neural models struggle to generalise in such evaluation setups.

There are also studies that consider compositional generalisation on more natural data.
Such studies typically focus on either MT \citep{lake2018generalization,raunak2019compositionality,li2021compositional} or semantic parsing \citep{finegan2018improving,keysers2019measuring,kim2020cogs,shaw-etal-2021-compositional}.
Most of these studies consider small and highly controlled subsets of natural language.

Instead, we focus on models trained on fully natural MT datasets, which we believe to be the setup for compositionality evaluation that does most justice to the complexity of natural language: contrary to semantic parsing, where the outputs are structures created by expert annotators, in translation both inputs and outputs are fully-fledged natural language sentences.
To the best of our knowledge, the only attempt to explicitly measure compositional generalisation of NMT models trained on large natural MT corpora is the study presented by \citet{raunak2019compositionality}.
They measure productivity -- generalisation to longer sentence lengths -- of an LSTM-based NMT model trained on a full-size, natural MT dataset.
Other studies using NMT, instead, consider toy datasets generated via templating \citep{lake2018generalization} or focus on short sentences excluding more complex constructions that contribute to the complexity of natural language for compositional generalisation, such as polysemous words or metaphors \citep{li2021compositional}.

%% file: discussion.tex
\section{Discussion}\label{sec:discussion}

Whether neural networks can generalise compositionally is often studied using artificial tasks that assume strictly \emph{local} interpretations of compositionality.
We argued that such interpretations exclude large parts of language and that to move towards human-like productive usage of language, tests are needed that assess how compositional models trained on \emph{natural data} are.\footnote{\citet{dupoux2018cognitive} makes a similar point for models of language acquisition, providing several concrete examples where using less than fully complex data proved problematic.}
We laid out reformulations of three compositional generalisation tests -- systematicity, substitutivity and overgeneralisation -- for NMT models trained on natural corpora, and assessed models trained on different amounts of data.
Our work provides an empirical contribution but also highlights vital hurdles to overcome when considering what it means for models of natural language to be compositional.
Below, we reflect on these hurdles and our results.

\paragraph{The proxy-to-meaning problem}
Compositionality is a property of the mapping between the form and meaning of an expression.
Since translation is a \emph{meaning-preserving} mapping from form in one language to form in another, it is an attractive task to evaluate compositionality: the translation of its sentence can be seen as a proxy to its meaning.
However, while expressions are assumed to have only one meaning, translation is a \emph{many-to-many} mapping: the same sentence can have multiple correct translations.
This does not only complicate evaluation -- MT systems are typically evaluated with BLEU because accuracy is not a suitable option -- it also raises questions about how compositional the desired behaviour of an MT model should be.
On the one hand, one could argue that for optimal generalisation, robustness, and accountability, we like models to behave systematically and consistently: we expect the translations of expressions to be independent of unrelated contextual changes that do not affect their meaning (e.g.\ swapping out a synonym in a nearby sentence).
Additionally, model performance could be improved if small changes do not introduce errors in unrelated parts of the translation.
On the other hand, non-compositional behaviour is not always incorrect -- it is one of the main arguments in our plea to test compositionality `in the wild' -- and we observe that indeed, not all non-compositional changes alter the correctness of the resulting translations.
Changing a translation from ``atleet'' (``athlete'') to ``sporter'' (``sportsman'') based on an unrelated word somewhat far away may not be (locally) compositional, but is it a problem?
And how do we separate such `harmful' mistakes from helpful ones?

\paragraph{The locality problem}
Inextricably linked to the proxy-to-meaning problem is the locality problem.
In our tests we see that \emph{small, local source changes} elicit \emph{global changes in translations}.
For instance, in our systematicity tests, changing one noun in a sentence elicited changes in the translation of a sentence that it was conjoined with.
In our substitutivity test, even synonyms that merely differed in spelling (e.g. ``doughnut'' and ``donut'') elicited changes to the remainder of the sentence.
This counters the idea of compositionality as a means of productively reusing language: if a phrase's translation depends on (unrelated) context that is not in its direct vicinity, this suggests that more evidence is required to acquire the translation of this phrase.

Tests involving synthetic data present the models with sentences in which maximally local behaviour is possible, and we argue that it is, therefore, also desirable.
Our experiments show that even in such setups, models do not translate in a local fashion: with varying degrees of correctness, they frequently change their translation when we slightly adapt the input.
On the one hand, this well-known \emph{volatility} \citep[see also][]{fadaee2020unreasonable} might be essential for coping with ambiguities for which meanings are context-dependent.
On the other hand, our manual analysis shows that the observed non-compositional behaviour does not reflect the incorporation of necessary contextual information and that oftentimes it is even altering the correctness of the translations.
Furthermore, this erratic behaviour highlights a lack of default reasoning, which can, in some cases, be problematic or even harmful, especially if faithfulness \citep{parthasarathi2021sometimes} or consistency is important.

In linguistics, it has been discussed how to extend the syntax and semantics such that `problem cases’ can be a part of a compositional language \citep{westerstaahl2002compositionality,pagin2010compositionality}.
In such formalisations, global information is used to disambiguate the problem cases, while other parts of the language are still treated locally.
In our models, global behaviour appears in situations where a local treatment would be perfectly suitable and where there is no clear evidence for ambiguity.
We follow \citet{baggio2021compositionality} in suggesting that we should learn from strategies employed by humans, who can assign compositional interpretations to expressions but can for some inputs also derive non-compositional meanings.
For \textit{human-like} linguistic generalisation, it is vital to investigate how models can represent both these types of processing, providing a locally compositional treatment when possible and deviating from that when needed.

\paragraph{Conclusion}
In conclusion, with this work, we contribute to the question of how compositional models trained on \emph{natural} data are, and we argue that MT is a suitable and relevant testing ground to ask this question.
Focusing on the balance between \emph{local} and \emph{global} forms of compositionality, we formulate three different compositionality tests and discuss the issues and considerations that come up when considering compositionality in the context of natural data.
Our tests indicate that models show both local and global processing, but not necessarily for the right samples.
Furthermore, they underscore the difficulty of separating helpful and harmful types of non-compositionality, stressing the need to rethink the evaluation of compositionality using natural language, where composing meaning is not as straightforward as doing the math.

%% file: acknowledgments.tex
\section*{Acknowledgements}

We thank Sebastian Riedel, Douwe Kiela, Thomas Wolf, Khalil Sima'an, Marzieh Fadaee, Marco Baroni, Brenden Lake and Adina Williams for providing feedback on this draft and our work in several different stages of it.
We thank Michiel van der Meer for contributing to the initial experiments that led to this paper.
A special thanks goes to Angela Fan, who assisted us at several points to get the ins and outs of training large MT models and double-checked several steps of our pipeline and to our ARR reviewers, who provided amazingly high quality feedback.
VD is supported by the UKRI Centre for Doctoral Training in Natural Language Processing, funded by the UKRI (grant EP/S022481/1) and the University of Edinburgh.

%% file: appendix.tex
\onecolumn
\section{Semi-natural templates}
\label{ap:ddop}

The semi-natural data that we use in our test sets is generated with the library \texttt{DiscoDOP},\footnote{\url{https://github.com/andreasvc/disco-dop}} developed for data-oriented parsing \citep{vancranenburgh2016disc}. 
We generate the data with the following seven step process:

\begin{enumerate}[wide, labelwidth=!, labelindent=0pt, itemsep=0pt, parsep=3pt]
\item[\textbf{Step 1.}] Sample 100k English OPUS sentences. 
\item[\textbf{Step 2.}] Generate a treebank using the disco-dop library and the \texttt{discodop parser en\_ptb} command. The library was developed for discontinuous data-oriented parsing. Use the library's \texttt{--fmt bracket} to turn off discontinuous parsing.
\item[\textbf{Step 3.}] Compute tree fragments from the resulting treebank (\texttt{discodop fragments}). These tree fragments are the building blocks of a Tree-Substitution Grammar.
\item[\textbf{Step 4.}] We assume the most frequent fragments to be common syntactic structures in English. To construct complex test sentences, we collect the 100 most frequent fragments containing at least 15 non-terminal nodes for NPs and VPs.
\item[\textbf{Step 5.}] Selection of three VP and five NP fragments to be used in our final semi-natural templates. These structures are selected through qualitative analysis for their diversity.
\item[\textbf{Step 6.}] Extract sentences matching the eight fragments (\texttt{discodop treesearch}).
\item[\textbf{Step 7.}] Create semi-natural sentences by varying one lexical item and varying the matching NPs and VPs retrieved in Step 6.
\end{enumerate}

In Table~\ref{tab:semi_natural_full}, we provide examples for each of the ten templates used, along with the internal structure of the complex NP or VP that is varied in the template.
In Table~\ref{tab:synthetic_data_appendix}, we provide some additional examples for our ten synthetic templates.

\begin{table}[!hb]
\centering\small
\begin{tabular}{lll}
\toprule
$n$& \textbf{Template} \\ \midrule\midrule
1  & The \bl{N}\SPSB{}{people} \textcolor{purple}{(VP (TO ) (VP (VB ) (NP (NP ) (PP (IN ) (NP (NP ) (PP (IN ) (NP )))))))} \\
   & \textit{E.g. The woman wants to use the Internet as a means of communication .}\\
2  & The \bl{N}\SPSB{}{people} \textcolor{purple}{(VP (VBP ) (VP (VBG ) (S (VP (TO ) (VP (VB ) (S (VP (TO ) (VP )))))))))} \\
   & \textit{E.g. The men are gon na have to move off-camera .}\\
3  & The \bl{N}\SPSB{}{people} \textcolor{purple}{(VP (VB ) (NP (NP ) (PP (IN ) (NP ))) (PP (IN ) (NP (NP ) (PP (IN ) (NP )))))} \\
   & \textit{E.g. The doctors retain 10 \% of these amounts by way of collection costs .} \\
4  & The \bl{N}\SPSB{}{people} reads an article about \textcolor{purple}{(NP (NP ) (PP (IN ) (NP (NP ) (PP (IN ) (NP (NP ) (PP (IN ) (NP )))))))} \\
   & \textit{E.g. The friend reads an article about the development of ascites in rats with liver cirrhosis .} \\
5  & The \bl{N}\SPSB{}{people} reads an article about \textcolor{purple}{(NP (NP (DT ) (NN )) (PP (IN ) (NP (NP ) (SBAR (S (WHNP (WDT )) (VP ))))))} . \\ 
   & \textit{E.g. The teachers read an article about the degree of progress that can be achieved by the industry .} \\
6  & An article about \textcolor{purple}{(NP (NP ) (PP (IN ) (NP (NP ) (PP (IN ) (NP (NP ) (PP (IN ) (NP )))))))} is read by the \bl{N}\SPSB{}{people} . \\
   & \textit{E.g. An article about the inland transport of dangerous goods from a variety of Member States is read by the lawyer .} \\
7  & An article about \textcolor{purple}{(NP (NP ) (PP (IN ) (NP (NP ) (, ,) (SBAR (S (WHNP (WDT )) (VP ))))))} , is read by the \bl{N}\SPSB{}{people} . \\
   & \textit{E.g. An article about the criterion on price stability , which was 27 \% , is read by the child .} \\
8 & Did the \bl{N}\SPSB{}{people} hear about \textcolor{purple}{(NP (NP ) (PP (IN ) (NP (NP ) (PP (IN ) (NP (NP ) (PP (IN ) (NP )))))))} . \\
   & \textit{E.g. Did the friend hear about an inhospitable fringe of land on the shores of the Dead Sea ?} \\
9 & Did the \bl{N}\SPSB{}{people} hear about \textcolor{purple}{(NP (NP (DT ) (NN )) (PP (IN ) (NP (NP ) (SBAR (S (WHNP (WDT )) (VP ))))))} ? \\
   & \textit{E.g. Did the teacher hear about the march on Employment which happened here on Sunday ?} \\
10 & Did the \bl{N}\SPSB{}{people} hear about \textcolor{purple}{(NP (NP ) (SBAR (S (VP (TO ) (VP (VB ) (NP (NP ) (PP (IN ) (NP ))))))))} ? \\
   & \textit{E.g. Did the lawyers hear about a qualification procedure to examine the suitability of the applicants ?}  \\
\bottomrule
\end{tabular}
\caption{Semi-natural data templates along with their identifiers ($n$). The syntactic structures for noun and verb phrases in purple are instantiated with data from the OPUS collection. 
Generated data from every template contains varying sentence structures and varying tokens but the predefined tokens in black remain the same.}
\label{tab:semi_natural_full}
\end{table}

\clearpage
\begin{table}[!h]
\small
\centering
\begin{tabular}{lll}
\toprule
$n$ & \textbf{Template} \\ \midrule\midrule
1   & The \bl{N}\SPSB{}{people} \bl{V}\SPSB{}{transitive} the \bl{N}\SPSB{sl}{people} . \\
    & \textit{E.g. The poet criticises the king .} \\
2   & The \bl{N}\SPSB{}{people} \bl{Adv} \bl{V}\SPSB{}{transitive} the \bl{N}\SPSB{sl}{people} . \\ 
    & \textit{E.g. The victim carefully observes the queen .}\\
3   & The \bl{N}\SPSB{}{people} \bl{P} the \bl{N}\SPSB{sl}{vehicle} \bl{V}\SPSB{}{transitive} the \bl{N}\SPSB{sl}{people} . \\ 
    & \textit{E.g. The athlete near the bike observes the leader .} \\
4   & The \bl{N}\SPSB{}{people} and the \bl{N}\SPSB{}{people} \bl{V}\SPSB{pl}{transitive} the \bl{N}\SPSB{sl}{people} . \\
    & \textit{E.g. The poet and the child understand the mayor .} \\
5   & The \bl{N}\SPSB{sl}{quantity} of \bl{N}\SPSB{pl}{people} \bl{P} the \bl{N}\SPSB{sl}{vehicle} \bl{V}\SPSB{sl}{transitive} the \bl{N}\SPSB{sl}{people} . \\
    & \textit{E.g. The group of friends beside the bike forgets the queen .} \\
6   & The \bl{N}\SPSB{}{people} \bl{V}\SPSB{}{transitive} that the \bl{N}\SPSB{pl}{people} \bl{V}\SPSB{pl}{intransitive}. \\
    & \textit{E.g. The farmer sees that the lawyers cry .} \\
7   & The \bl{N}\SPSB{}{people} \bl{Adv} \bl{V}\SPSB{}{transitive} that the \bl{N}\SPSB{pl}{people} \bl{V}\SPSB{pl}{intransitive} . \\
    & \textit{E.g. The mother probably thinks that the fathers scream .} \\
8   & The \bl{N}\SPSB{}{people} \bl{V}\SPSB{}{transitive} that the \bl{N}\SPSB{pl}{people} \bl{V}\SPSB{pl}{intransitive} \bl{Adv} . \\
    & \textit{E.g. The mother thinks that the fathers scream carefully .} \\
9   & The \bl{N}\SPSB{}{people} that \bl{V}\SPSB{}{intransitive} \bl{V}\SPSB{}{transitive} the \bl{N}\SPSB{sl}{people} . \\
    & \textit{E.g. The poets that sleep understand the queen .} \\
10  & The \bl{N}\SPSB{}{people} that \bl{V}\SPSB{}{transitive} \bl{Pro} \bl{V}\SPSB{sl}{transitive} the \bl{N}\SPSB{sl}{people} . \\
    & \textit{E.g. The mother that criticises him recognises the queen .} \\
    \bottomrule
    \end{tabular}
    \captionof{table}{Synthetic sentence templates similar to \citet{lakretz2019emergence}, along with their identifiers ($n$). }
    \label{tab:synthetic_data_appendix}
\vspace{-0.3cm}
\end{table}

\section{Systematicity}
\label{ap:systematicity}

Table~\ref{tab:systematicity_appendix} provides the numerical counterparts of the results visualised in Figure~\ref{fig:systematicity}.

\begin{table*}[!h]
    \centering\small\setlength{\tabcolsep}{4pt}
    \begin{subtable}[b]{0.49\textwidth}
    \begin{tabular}{lcccc}
    \toprule
    \textbf{Data} &     \textbf{Condition} & \multicolumn{3}{c}{\textbf{Model}} \\
    & & small & medium & full \\
    \midrule\midrule
    \texttt{S\;$\rightarrow$\;NP\;VP} \\
    synthetic & NP & .73 & .84 & .84 \\
    synthetic & VP & .76 & .87 & .88 \\
    semi-natural & NP & .63 & .66 & .64 \\ \midrule
    \texttt{S\;$\rightarrow$\;S\;CONJ\;S} \\
    synthetic & $\text{S}^\prime_1$ & .81 & .90 & .92 \\
    synthetic & $\text{S}_3$ & .53 & .76 & .82 \\
    semi-natural & $\text{S}^\prime_1$ & .65 & .73 & .76 \\
    semi-natural & $\text{S}_3$ & .29 & .49 & .49 \\
    natural & $\text{S}^\prime_1$ & .58 & .67 & .72 \\
    natural & $\text{S}_3$ & .25 & .39 & .47 \\
    \bottomrule
    \end{tabular}
    \caption{Per models' training set size}
    \end{subtable}
    \begin{subtable}[b]{0.49\textwidth}
    \centering\small\setlength{\tabcolsep}{4pt}
    \begin{tabular}{cccccccccc}
    \toprule
    \multicolumn{10}{c}{\textbf{Template}} \\
    1 & 2 & 3 & 4 & 5 & 6 & 7 & 8 & 9 & 10 \\
    \midrule\midrule
    \\
    .86 & .74 & .85 & .87 & .75 & .89 & .85 & .85 & .70 & .68 \\
    .92 & .73 & .90 & .91 & .84 & .88 & .85 & .82 & .77 & .74 \\
    .66 & .63 & .65 & .70 & .64 & .69 & .63 & .63 & .60 & .58 \\ \midrule \\
    .91 & .82 & .88 & .88 & .86 & .95 & .90 & .91 & .84 & .79 \\
    .75 & .54 & .72 & .66 & .73 & .88 & .74 & .81 & .66 & .55 \\
    .73 & .75 & .75 & .80 & .75 & .73 & .66 & .68 & .64 & .64 \\
    .50 & .50 & .51 & .58 & .52 & .43 & .35 & .31 & .28 & .29 \\
    .67 & .74 & .65 & .64 & .63 & .64 & .62 & .66 & .63 & .66 \\
    .39 & .49 & .35 & .35 & .34 & .37 & .33 & .38 & .34 & .38 \\
    \bottomrule
    \end{tabular}
    \caption{Per template}
    \end{subtable}
    \caption{Consistency scores for the systematicity experiments, detailed per experimental setup and evaluation data type. We provide scores (a) per models' training set size, and (b) per template of our generated evaluation data. For natural data, the template number is meaningless, apart from the fact that it determines sentence length and word frequency.}
    \label{tab:systematicity_appendix}
\end{table*}

\section{Substitutivity}
\label{ap:substitutivity}

\paragraph{Synonyms employed} In Table~\ref{tab:freqs_substitutivity_appendix}, we provide some information about the synonymous word pairs used in the substitutivity test, including their frequency in OPUS and their most common Dutch translation. 
The last column of the table contains the subordinate clauses that we used to include the synonyms in the synthetic and semi-natural data. 
We include them as a relative clause behind nouns representing a human, such as ``The poet criticises the king that eats the doughnut''.

\paragraph{Detecting synonym translations}
To find the span of text in the translation which is the translation of the synonym, we apply a relatively simple heuristic.
We generate a number of short sentences such as ``This is the NOUN'', feed those to all our trained models, and extract the top-5 answers in the beam.
We then use the list of all words resulting from this protocol -- which we manually checked -- to find synonym translations in the model output.

\paragraph{Results} In the main paper, Figures~\ref{fig:substitutivity} and~\ref{fig:per_synonym} provided the consistency scores for the substitutivity tests.
Here, Table~\ref{tab:substitutivity_appendix} further details the results from the figure, by presenting the average consistency per evaluation data type and training set size, and per evaluation data type and synonym pair.

\noindent\begin{minipage}[b]{0.99\textwidth}
\vspace{1cm}\small\centering
\begin{tabular}{llllll}
    \toprule
    \multicolumn{4}{l}{\textbf{Synonym pair}} & \textbf{Dutch translation} & \textbf{Subordinate clause} \\
    \textit{British} & \textit{Freq.} & \textit{American} & \textit{Freq.} \\\midrule \midrule
    aeroplane & 6728 & airplane & 5403 & vliegtuig & that travels by \dots \\
    aluminium & 17982 & aluminum & 5700 & aluminium & that sells \dots \\
    doughnut & 2014 & donut & 1889 & donut & that eats the \dots \\
    foetus & 1943 & fetus & 1878 & foetus & that researches the \dots \\
    flautist & 112 & flutist & 101 & fluitist & that knows the \dots  \\
    moustache & 1132 & mustache & 1639 & snor & that has a \dots \\
    tumour & 7338 & tumor & 6348 & tumor & that has a \dots \\
    pyjamas & 808 & pajamas & 1106 & pyjama & that wears \dots \\
    sulphate & 3776 & sulfate & 1143 & zwavel & that sells \dots  \\
    yoghurt & 1467 & yogurt & 2070 & yoghurt & that eats the \dots \\
    aubergine & 765 & eggplant & 762 & aubergine & that eats the \dots \\
    shopping trolley & 217 & shopping cart & 13366 & winkelwagen & that uses a \dots \\
    veterinary surgeon & 941 & veterinarian & 6995 & dierenarts & that knows the \dots \\
    sailing boat & 5097 & sailboat & 1977 & zeilboot & that owns a \dots \\
    football & 33125 & soccer & 6841 & voetbal & that plays \dots \\
    holiday & 125430 & vacation & 23532 & vakantie & that enjoys the \dots \\
    ladybird & 235 & ladybug & 303 & lieveheersbeestje & that caught a \dots \\
    theatre & 19451 & theater & 13508 & theater & that loves \dots \\
    postcode & 479 & zip code & 1392 & postcode & with the same \dots \\
    whisky & 3604 & whiskey & 4313 & whisky & that drinks \dots \\
    \bottomrule
\end{tabular}
\captionof{table}{Synonyms for the substitutivity test, along with their OPUS frequency, Dutch translation, and the subordinate clause used to insert them in the data.}
\label{tab:freqs_substitutivity_appendix}
\end{minipage}

\begin{table*}[!h]
    \centering\small\setlength{\tabcolsep}{4pt}
    \begin{subtable}[b]{\textwidth}\centering
    \begin{tabular}{llccc}
    \toprule
    \textbf{Data} &     \textbf{Metric} & \multicolumn{3}{c}{\textbf{Model}} \\
    & & small & medium & full \\
    \midrule\midrule
    synthetic    & con.       & .49  & .67  & .76  \\
                 & syn. con.  & .67  & .82  & .93  \\
    semi-natural & con.       & .34  & .55  & .62  \\
                 & syn. con.  & .62  & .84  & .93  \\
    natural      & con.       & .37  & .52  & .63  \\
                 & syn. con.  & .61  & .75  & .85  \\    \bottomrule
    \end{tabular}
    \caption{Per models' training set size}
    \end{subtable}
    \begin{subtable}[b]{\textwidth}\setlength{\tabcolsep}{3pt}\centering
    \begin{tabular}{llcccccccccccccccccccc}
    \toprule
    \textbf{Data} &     \textbf{Metric} & \multicolumn{20}{c}{\textbf{Synonym}} \\
& & \rotatebox{90}{aeroplane} & \rotatebox{90}{aluminium} & \rotatebox{90}{doughnut} & \rotatebox{90}{foetus} & \rotatebox{90}{flautist} & \rotatebox{90}{moustache} & \rotatebox{90}{tumour} & \rotatebox{90}{pyjamas} & \rotatebox{90}{sulphate} & \rotatebox{90}{yoghurt} & \rotatebox{90}{aubergine} & \rotatebox{90}{shopping trolley} & \rotatebox{90}{veterinary surgeon} & \rotatebox{90}{sailing boat} & \rotatebox{90}{football} & \rotatebox{90}{holiday} & \rotatebox{90}{ladybird} & \rotatebox{90}{theatre} & \rotatebox{90}{postcode} & \rotatebox{90}{whisky} \\
    \midrule\midrule
synthetic & con.    & .54  & .87  & .74  & .82  & .10  & .92  & .78  & .64  & .79  & .55  & .25  & .40  & .64  & .73  & .68  & .81  & .27  & .85  & .48  & .88  \\
 & syn. con.        & 1.0  & 1.0  & .87  & 1.0  & .10  & 1.0  & 1.0  & .80  & .95  & 1.0  & .38  & .48  & .90  & 1.0  & .75  & 1.0  & .40  & .99  & .53  & 1.0  \\
semi-natural & con. & .43  & .59  & .58  & .54  & .08  & .85  & .52  & .55  & .56  & .42  & .24  & .31  & .33  & .73  & .66  & .71  & .20  & .62  & .43  & .75  \\
 & syn. con.        & .99  & .99  & .83  & 1.0  & .09  & 1.0  & .98  & .72  & .90  & .98  & .40  & .50  & .77  & 1.0  & .90  & 1.0  & .38  & .95  & .58  & .99  \\
natural & con.      & .50  & .52  & .53  & .56  & .09  & .75  & .50  & .60  & .47  & .57  & .23  & .70  & .29  & .64  & .55  & .62  & .17  & .59  & .61  & .58  \\
 & syn. con.        & .89  & .85  & .73  & .91  & .11  & .87  & .87  & .82  & .88  & .86  & .32  & .92  & .75  & .71  & .79  & .81  & .27  & .82  & .81  & .80  \\    \bottomrule
    \end{tabular}
    \caption{Per synonym}
    \end{subtable}
    \caption{Consistency scores for the substitutivity experiments, detailed per evaluation data type. We present scores (a) per models' training set size and (b) per synonym.}
    \label{tab:substitutivity_appendix}
\end{table*}

\clearpage

\section{Global compositionality}
\label{ap:global_compositionality}

\paragraph{Idioms employed}
Table~\ref{tab:overgeneralisation_appendix} provides more information on the idioms used in our global compositionality test. 
In the first column, we list all idioms we used, along with the \emph{keywords} that we used to determine if their translation is local or not.
To extract the natural data, we retrieved exact matches with OPUS source sentences. 
The idioms' keywords are mostly nouns that either translate into a different word in an accurate paraphrased translation in Dutch (e.g. ``across the \textbf{board}'' would be ``over de hele linie''), or should disappear in the translation (e.g. ``do the right \textbf{thing}'' typically translates into ``het juiste doen'' in the corpus). 

In the second column of Table~\ref{tab:overgeneralisation_appendix}, we list the subordinate clauses that we used to include idioms in the synthetic and semi-natural data.
The clauses themselves are drawn from source sentences in OPUS.
To incorporate them in synthetic and semi-natural sentences, we include them as a relative clause behind nouns representing a human, by attaching \emph{``that said `[idiom]'''}. 
For instance: ``The poet criticises the king that said `Have you gone out of your mind'.''

In the third column of Table~\ref{tab:overgeneralisation_appendix}, we show local translations of the idioms, elicited from the model by embedding the idiom in a string of ten random nouns. 
Even ``out of the blue'', which is rarely overgeneralised when presented in synthetic, semi-natural or natural contexts, is locally translated. 
This indicates that the idiom is not stored as one lexical unit per se but that it is only translated globally in specific contexts.

\paragraph{Results}
In the main paper, in Figure~\ref{fig:global_compositionality}, we visualised how overgeneralisation changes over the course of training, averaged over idioms.
In Table~\ref{tab:global_compositionality_appendix}, we detail the maximum overgeneralisation observed per idiom.

\vspace{1cm}
\noindent\begin{minipage}[t]{\textwidth}
\small
\centering
\resizebox{0.8\textwidth}{!}{\begin{tabular}{lll}
\toprule
\textbf{Idiom} & \textbf{Subordinate clause} & \textbf{Local translation} \\ \midrule\midrule
\underline{once} in a \underline{while} & that said `` I will play it once in a while " & eens in een tijdje \\
do the right \underline{thing} & that said `` Just do the right thing " & doen het juiste ding \\
out of your \underline{mind} & that said `` Have you gone out of your mind " & uit je hoofd \\
\underline{state} of the \underline{art} & that said `` This is a state of the art, official facility " & stand van de kunst \\
from \underline{scratch} & that said `` We are cooking from scratch every day " & van kras\\
take \underline{stock} & that said `` Take stock of the lessons to be drawn " & nemen voorraad  \\
across the \underline{board} & that said `` I got red lights all across the board " & aan boord\\
in the final \underline{analysis} & that said `` In the final analysis, this is what matters " & in de laatste analyse\\
out of the \underline{blue} & that said `` It just came out of the blue " & uit het blauwe \\
in \underline{tandem} & that said `` We will work with them in tandem " & in tandem \\
by \underline{heart} & that said `` I knew the formula by heart " & door hart \\
come to \underline{terms} with & that said `` I have come to terms with my evil past " & komen overeen met \\
by the same \underline{token} & that said `` By the same token I will oppose what is evil " & bij dezelfde token \\
at your \underline{fingertips} & that said `` The answer is right at your fingertips " & binnen handbereik \\
look the other \underline{way} & that said `` We cannot look the other way either " & kijken de andere manier \\
follow \underline{suit} & that said `` And many others follow suit " & volgen pak \\
keep \underline{tabs} on & that said `` I keep tabs on you " & houden tabs \\
in the short \underline{run} & that said `` In the short run it clearly must be " & in de korte lopen \\
by \underline{dint} of & that said `` We are part of it by dint of our commitment " & door de int \\
set \underline{eyes} on & that said `` I wish I had never set eyes on him " & set ogen op \\

\bottomrule
\end{tabular}}
\captionof{table}{Idioms used in the overgeneralisation test. 
    The words that are indicative of a local translation are underlined, we check for their presence to label a translation as an overgeneralisation.
    The listed subordinate clauses are used to insert the idioms into synthetic and semi-natural templates.
    The local translation indicated is the translation given by the model when the idiom is embedded in a string of ten random words.
    }
\label{tab:overgeneralisation_appendix}
\end{minipage}

\begin{table*}[!ht]
    \centering\small\setlength{\tabcolsep}{3.5pt}
    \begin{tabular}{llcccccccccccccccccccc}
    \toprule
    \textbf{Data} & \textbf{Model} & \multicolumn{20}{c}{\textbf{Idiom}} \\
    & & \rotatebox{90}{once in a while} &
        \rotatebox{90}{do the right thing} & 
        \rotatebox{90}{out of your mind} & 
        \rotatebox{90}{state of the art} &
        \rotatebox{90}{from scratch} & 
        \rotatebox{90}{take stock} & 
        \rotatebox{90}{across the board} & 
        \rotatebox{90}{in the final analysis} &
        \rotatebox{90}{out of the blue} &
        \rotatebox{90}{in tandem} &
        \rotatebox{90}{by heart} & 
        \rotatebox{90}{come to terms with} & 
        \rotatebox{90}{by the same token} & 
        \rotatebox{90}{look the other way} & 
        \rotatebox{90}{at your fingertips} & 
        \rotatebox{90}{follow suit} & 
        \rotatebox{90}{keep tabs on} & 
        \rotatebox{90}{in the short run} & 
        \rotatebox{90}{by dint of} & 
        \rotatebox{90}{set eyes on} \\
    \midrule \midrule
    synthetic & small   & .98 & .92 & .98 & 1.0 & .40 & .75 & 1.0 & 1.0 & 1.0 & 1.0 & 1.0 & .01 & 1.0 & 1.0 & 1.0 & .99 & 1.0 & .72 & .20 & .74 \\
     & medium           & .99 & .96 & .98 & 1.0 & .76 & .73 & 1.0 & 1.0 & 1.0 & 1.0 & 1.0 & .22 & 1.0 & 1.0 & 1.0 & 1.0 & .57 & .55 & .38 & .57 \\
     & full             & .97 & .86 & .97 & 1.0 & .50 & .56 & 1.0 & 1.0 & 1.0 & 1.0 & 1.0 & .24 & 1.0 & .91 & 1.0 & 1.0 & .74 & .38 & .24 & .44 \\
    semi-natural & small& .95 & .66 & .98 & 1.0 & .49 & .73 & 1.0 & 1.0 & 1.0 & .97 & 1.0 & .08 & 1.0 & .98 & 1.0 & .88 & .99 & .56 & .15 & .81 \\
     & medium           & .91 & .60 & .95 & 1.0 & .78 & .63 & .96 & 1.0 & 1.0 & .97 & 1.0 & .31 & .99 & .99 & 1.0 & .97 & .74 & .45 & .30 & .59 \\
     & full             & .97 & .55 & .95 & 1.0 & .40 & .68 & .99 & 1.0 & 1.0 & .99 & 1.0 & .31 & 1.0 & .90 & 1.0 & .97 & .90 & .25 & .23 & .47 \\
    natural & small     & .80 & .51 & .80 & .97 & .84 & .31 & .75 & .96 & .92 & .82 & .88 & .14 & .74 & .60 & 1.0 & .40 & .96 & .29 & .23 & .87 \\
     & medium           & .80 & .50 & .82 & .96 & .84 & .32 & .71 & .94 & .92 & .68 & .90 & .22 & .74 & .63 & .99 & .39 & .61 & .33 & .29 & .84 \\
     & full             & .79 & .39 & .83 & .95 & .90 & .36 & .83 & .98 & .95 & .89 & .90 & .11 & .65 & .55 & 1.0 & .65 & .56 & .19 & .27 & .76 \\
 \bottomrule
    \end{tabular}
    \caption{Maximum overgeneralisation observed over the course of training, per evaluation data type, training set size and idiom.}
    \label{tab:global_compositionality_appendix}
\end{table*}

\section{Reproducibility details}\label{app:reproducibility}

\subsection{Data}

\paragraph{Training data} Our training data consists of the English-Dutch subset of the MT corpus \textsc{OPUS} \citep{tiedemann2020opus}, provided by \citet{tiedemann-2020-tatoeba}.
This data contains in total 69M source-target pairs.
The data can be found on
\url{https://github.com/Helsinki-NLP/Tatoeba-Challenge/blob/master/data/README-v2020-07-28.md}.

\paragraph{Preprocessing}
We tokenise the data using the tokenisation script\footnote{\url{https://github.com/moses-smt/mosesdecoder/blob/master/scripts/tokenizer/tokenizer.perl}} from the SMT library Moses.\footnote{\url{https://github.com/moses-smt/mosesdecoder}}
Following the number of subwords suggested by \citet{tiedemann-2020-tatoeba}, we generate a subword vocabulary applying 60k BPE merge-operations.
To do so, we use the \texttt{learn\_bpe.py} script provided in the \textsc{subword\_nmt}\footnote{\url{https://github.com/rsennrich/subword-nmt/blob/master/subword_nmt/learn_bpe.py}} repository hosted by Rico Sennrich.

\paragraph{Different corpora}
We train models on three different sizes of corpora: \textsc{small}, \textsc{medium} and \textsc{full}.
To generate these corpora, we first shuffle the OPUS training data using the bash function \texttt{shuffle}.
To generate the \textsc{small} and \textsc{medium} corpora, we take the first 8582811 and 1072851 sentences of this shuffled corpus, which corresponds to $\frac{1}{8}$th and $\frac{1}{64}$th of the full training corpus, respectively.
For each setting, we train models with seeds \{1, 2, 3, 4, 5\}.

\paragraph{Test and validation data}
Initially, we aimed to evaluate our models using the commonly used MT test sets OPUS-100\footnote{\url{http://data.statmt.org/opus-100-corpus/v1.0/supervised/en-nl/}} and the test partition of the TED talk corpus.\footnote{\url{https://github.com/neulab/word-embeddings-for-nmt}}
However, it turned out that both these test sets were almost fully contained in our training corpus.
We, therefore, adopted the newer \textsc{Flores-101} corpus \citep{goyal2021flores}, of which we used both the `dev' and the `devtest' set.
The data can be downloaded from \url{https://dl.fbaipublicfiles.com/flores101/dataset/flores101_dataset.tar.gz}.
To compute BLEU scores, we tokenised the data with the Moses tokenisation script mentioned above, and then used the commandline script \texttt{fairseq-generate} to compute scores.

We furthermore use several evaluation sets to assess the compositional abilities of our trained models.
The data for these tests, as well as scripts to run them and plot their results, can be found in the following repository: \url{https://github.com/i-machine-think/compositionality_paradox_mt}.

\subsection{Architecture and training} 
As reported in the main text, we focus on English-Dutch translation, and all our models are Transformer-base models, as implemented in Fairseq \citep{ott2019fairseq}.\footnote{We used the implementation as it was on May 12, 2021: \url{https://github.com/pytorch/fairseq/blob/d151f2787240cca4e3c7e47640e647f8ae028c37/fairseq/models/transformer.py}}
Both the encoder and the decoder of this model have an embedding dimension of 512, 6 layers, 8 attention heads and a feed-forward layer dimension of 2048.
With our vocabulary, the models have a total of around 80M trainable parameters.

To train our models, we follow the training procedure suggested by \citet{ott2018scaling}, which can be found at \url{https://github.com/pytorch/fairseq/tree/master/examples/scaling_nmt}.
To summarise, we share all embeddings between the encoder and the decoder, use Adam as optimiser with $\beta$-values (0.9, 0.98), starting from an initial warmup learning rate of 1e-07 for 4000 warmup updates and a learning rate of 0.0005 afterwards, using inverse square root as the learning rate scheduler.
We use a clip-norm of 0.0, dropout of 0.3, weight-decay of 0.0001, label-smoothing of 0.1.
The maximum number of tokens in a batch is 3584, we simulate larger batches by increasing the update frequency to 8.
To determine early stopping, we use a patience of 10 (i.e.\ we stop training if a model does not improve on the dev set anymore for 10 epochs, and take the best checkpoint at that point).
Any other hyperparameters involved follow the Fairseq default.
We provide the BLEU scores per model seed in Table~\ref{tab:ap_bleu}.

\subsection{Compute}
All experiments were ran using Tesla V100 GPUs on an internal SLURM-based cluster. 
Training a transformer-base model on our small, medium and full dataset takes on average 3.5, 17 and 113 minutes per epoch, respectively (numbers are rounded) on 32 GPUs.
This makes the total training time for these models, which are trained for around 160, 60 and 30 epochs, 10,  17 and 56 hours, respectively (again, spread over 32 GPUs).

\begin{table}\centering\small
\begin{tabular}{lccc}
\toprule
\textbf{Training set size} & \textbf{Seed} & \textbf{BLEU dev} & \textbf{BLEU devtest} \\ \midrule \midrule
small & 1 & 20.92 & 21.14 \\
      & 2 & 20.77 & 20.37 \\
      & 3 & 20.42 & 20.11 \\
      & 4 & 20.95 & 20.23 \\
      & 5 & 20.88 & 20.84 \\\midrule
medium& 1 & 24.09 & 24.18 \\
      & 2 & 25.05 & 24.71 \\
      & 3 & 24.55 & 24.42 \\
      & 4 & 24.09 & 23.93 \\
      & 5 & 24.55 & 24.10 \\\midrule
full  & 1 & 26.17 & 25.63 \\
      & 2 & 25.71 & 25.63 \\
      & 3 & 25.82 & 25.72 \\
      & 4 & 26.19 & 25.84 \\
      & 5 & 25.86 & 25.76 \\\bottomrule
\end{tabular}
\caption{BLEU scores for the `dev' and `devtest' subsets of the \textsc{Flores} datasets, for models trained on corpora of three sizes, for five seeds per training set size.}
\label{tab:ap_bleu}
\end{table}

%% file: appendix_analysis.tex
\onecolumn
\section{Manual analysis}\label{app:man_analysis}

Our quantitative tests provide information on when a model behaves locally and when globally in automated form but they do not consider whether that behaviour is incorrect or not.
More simply put, we do not know whether the changes that we observe are actually resulting in incorrect translations.
We complement these scores with an elaborate manual analysis, which provides more insight into the nature of the non-compositional behaviour we registered.

\subsection{Setup}

\paragraph{Data sampling} We randomly sample 900 examples for substitutivity (100 for each \{model\}$\times$\{test data type\} tuple) and 900 examples for systematicity (50 for each \{model\}$\times$\{test data type\}$\times$\{S$_1^\prime$, S$_3$\} tuple), randomly distributed over templates.
In all cases, we sample sentences randomly from the five seeds that we trained, and from all templates.
For substitutivity, we sample five examples for each synonym for every \{model\}$\times$\{test data type\} pair.

\paragraph{Annotation procedure}
For each of these samples, we annotate how they differ, where we distinguish between four general categories:
\begin{itemize}[noitemsep,topsep=0pt]
\item[i.] \emph{Rephrasing}: part of the sentence is rephrased (but both phrases are equally (in)correct);
\item[ii.] \emph{Source ambiguities}: there is an ambiguity in the source sentence, and the model switches its interpretation; 
\item[iii.] \emph{Errors}: one of the translations contains an error that the other one does not;
\item[iv.] \emph{Formatting}: minor formatting changes, consisting mostly of insertions/deletions of punctuation.
\end{itemize}

\noindent For the substitutivity data, we separately annotate changes that are related to the translation of the synonym, where we distinguish cases in which both synonyms are correctly or incorrectly translated from cases in which one of the translations is correct.
We annotate all changes observed in a sample -- one sentence may thus contain annotations for multiple changes -- and report the relative frequency of each class of errors.

\begin{figure}[h]\centering
\begin{subfigure}[b]{0.25\textwidth}
\includegraphics[width=\textwidth]{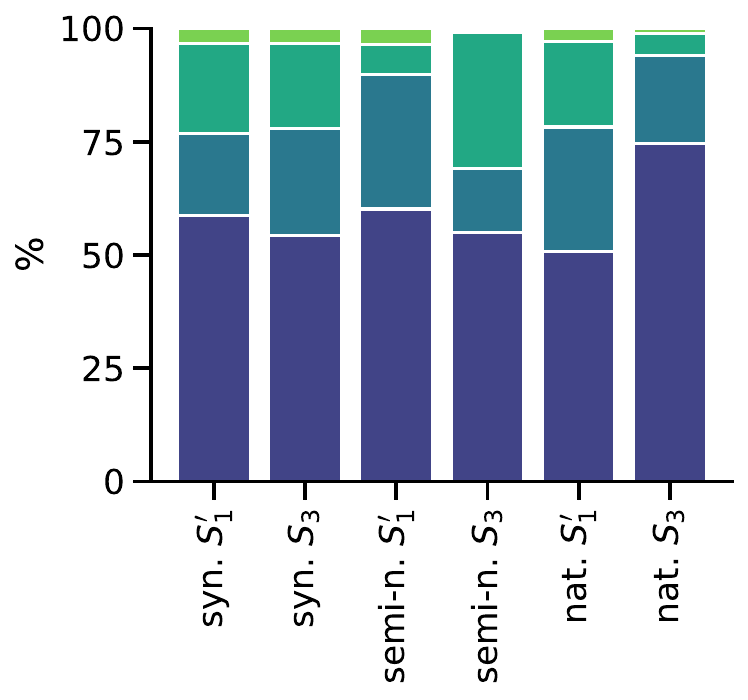}
\caption{Small training set}
\end{subfigure}
\begin{subfigure}[b]{0.25\textwidth}
\includegraphics[width=\textwidth]{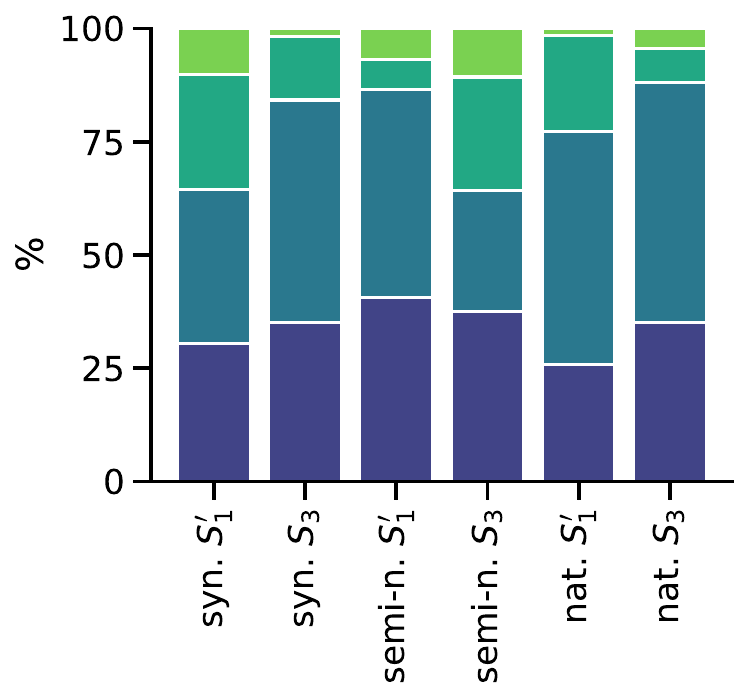}
\caption{Medium training set}
\end{subfigure}
\begin{subfigure}[b]{0.41\textwidth}
\includegraphics[width=0.9\textwidth]{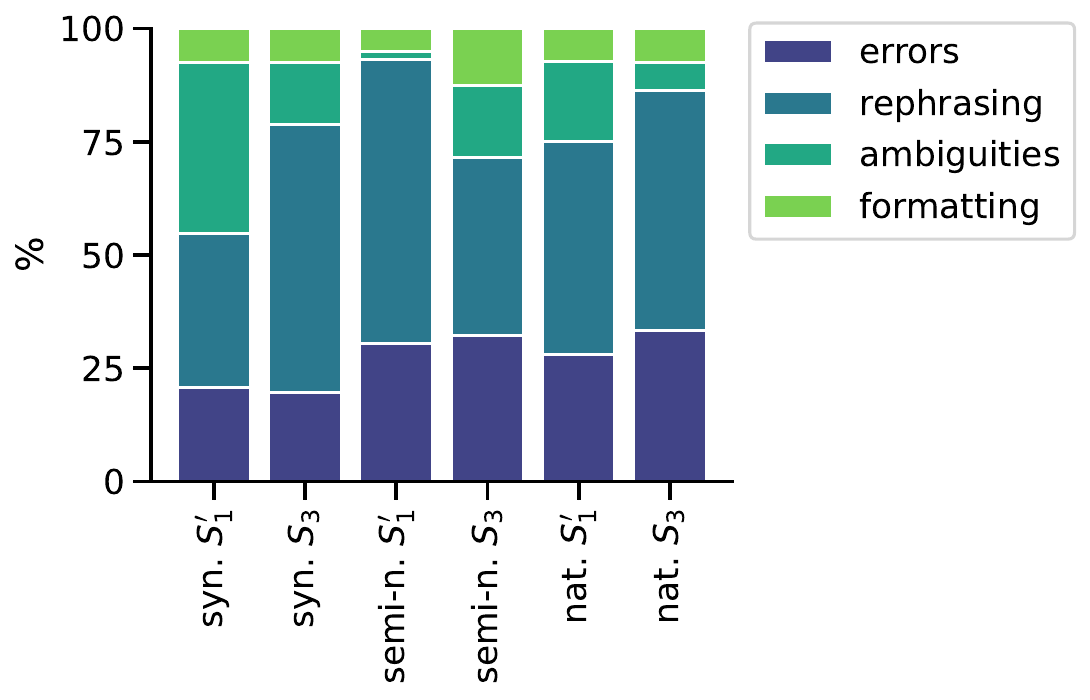}
\caption{Full training set}
\end{subfigure}
\caption{Distribution of error types for sentences that contain inconsistencies in systematicity, detailed per model trained on the training set sizes in the subcaptions.}
\label{fig:ap_systematicity_analysis}
\end{figure}
\begin{figure}[t]\centering
\begin{subfigure}[b]{0.22\textwidth}\centering
\includegraphics[width=0.79\textwidth]{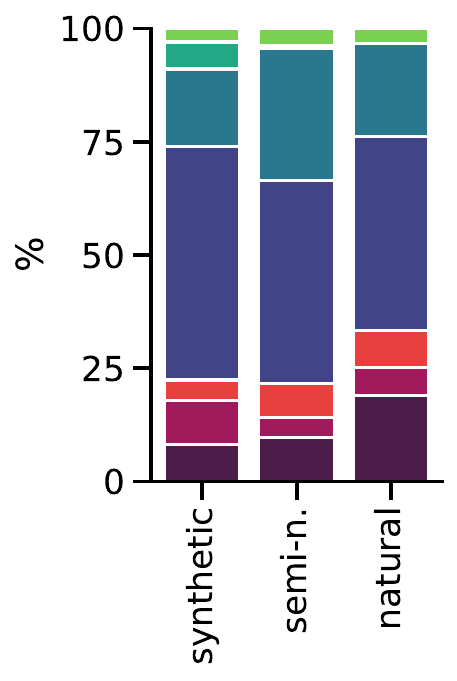}
\caption{Small training set}
\end{subfigure}
\begin{subfigure}[b]{0.22\textwidth}\centering
\includegraphics[width=0.79\textwidth]{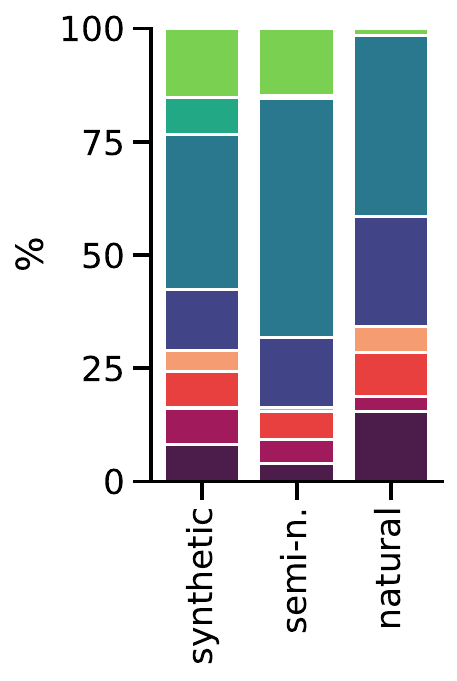}
\caption{Medium training set}
\end{subfigure}
\begin{subfigure}[b]{0.43\textwidth}\centering
\includegraphics[width=0.85\textwidth]{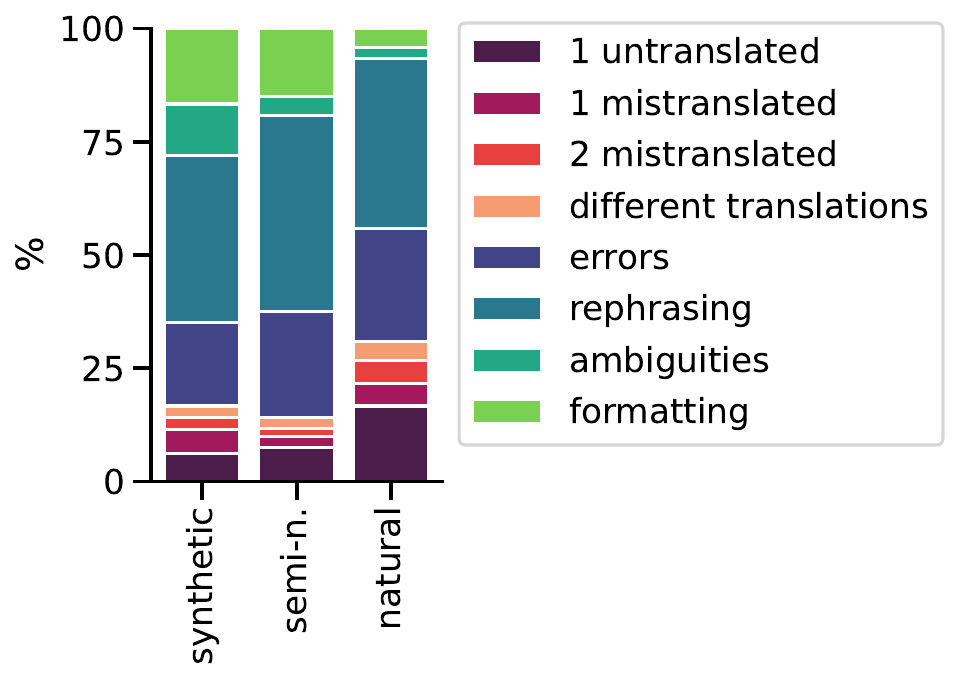}
\caption{Full training set}
\end{subfigure}
\caption{Distribution of the types of inconsistencies observed in the substitutivity test, detailed per model trained on the training set sizes in the subcaptions.
The red colour scheme represents error types specific to this experiment.}
\label{fig:ap_substitutivity_analysis}
\vspace{-.3cm}
\end{figure}

\subsection{Results}

We provide a summary of the results in Figure~\ref{fig:ap_systematicity_analysis} for systematicity and Figure~\ref{fig:ap_substitutivity_analysis} for substitutivity.
As a general trend, the results reflect that in models trained on smaller datasets, more mistakes are actually errors, rather than multiple correct alternatives.
In the systematicity test, 59\% of the inconsistencies for the models trained on the smallest dataset are erroneous changes, versus 34\% and 27\% in the models trained on the medium and largest dataset,
when we average the percentages over the different subsets annotated.
For substitutivity, the percentage of erroneous changes unrelated to the synonyms comprises 46\%, 18\% and 22\% for the smallest, medium and full dataset, respectively.
On top of that, there were inconsistencies related to the synonyms, that represented 26\%, 26\% and 21\% for the three dataset sizes, respectively.
While this is expected, to some extent, it still constitutes a problem: for models trained on smaller amounts of data, being able to translate in a compositional manner is particularly relevant.
Below, we further elaborate on the types of inconsistencies encountered per annotation category, including some examples.

\subsubsection{Rephrasing}
A large portion of the inconsistencies concerns pairs where one translation can be considered a rephrased version of the other translation.
A common cause of this is a \textbf{reordering of words} that does not impact the grammaticality or meaning of the Dutch sentence -- e.g.\ in sentences with adverbs (\exa{heeft de burgemeester zeker in de gaten} vs \exa{heeft zeker de burgemeester in de gaten}) or relative clauses with direct objects (\exa{die genieten van de vakantie} vs \exa{die van de vakantie genieten}).
We could not trace these reorderings back to the specific change made in the systematicity or substitutivity tests.
Consider, for instance, Example \ref{ex:zeker}, where the reordering happens as a consequence of changing the word \exa{king} to \exa{father}.
Note also that while these translations both contain an error (\exa{neemt \dots in de gaten}), this is not marked as an inconsistency, because it is shared between the translations.

\ex.\label{ex:zeker}
\a. \textsc{EN}: The aunts criticise the \{king, father\}, and the man definitely observes the mayor.
\b. \textsc{NL}: (\dots) en de man neemt zeker de burgemeester in de gaten.
\c. \textsc{NL}: (\dots) en de man neemt de burgemeester zeker in de gaten.

Another commonly occurring case of rephrasing is one where the two translations include terms that are (nearly) \textbf{synonymous terms} in Dutch.
Some examples are the translation of athlete (\exa{sporter} vs \exa{atleet}), wish (\exa{wensen} vs \exa{willen}) and observe (\exa{observeren} vs \exa{waarnemen}).
Some of them can appear in the same context but for others the two words would typically appear in different types of texts.
For instance, the word \exa{dokter} is used in more informal contexts than the word \exa{arts} (both translations of \exa{doctor}).
Again, we could not identify an interpretable pattern for when the model emits one instead of the other -- they were not understandably related to the modifications we made to the inputs.

\subsubsection{Source ambiguities}
An intriguing category that we had not anticipated were cases in which the source sentence contained ambiguities, such as \textbf{polysemous words} 
(e.g.\ \exa{director} translated to \exa{directeur}, referring to the director of a company, and \exa{regisseur}, indicating the director of a movie).
Other ambiguities encountered were \textbf{scope ambiguities}, that were particularly prominent for the systematicity test.
In that test, we concatenate two sentences, and the ambiguity was often related to the verb in the first sentence -- e.g.\ in Example~\ref{ex:director}: 

\ex.\label{ex:director}
\a. \textsc{EN}: The friend wishes that the \{lawyers, directors\} scream, and the victims (\dots)

While we intended this to be a conjunction of two independent sentences, there is also a reading where \exa{wishes} takes scope over the entire second conjunct.
In Dutch, those two cases are distinguishable because they trigger a different word order in the embedded clause (SOV), which is not grammatical for main clauses.
Such scope changes often lead to very questionable interpretations of the English sentence, as is the case for Example~\ref{ex:2CV}:

\ex.\label{ex:2CV}
\a. \textsc{EN}: The victims want that the \{doctors, mayors\} run, and the victims read an article about the case of a procedure which includes a repayment plan.
\b. \textsc{EN}: The farmers think that the \{butchers, mothers\} laugh, and an error can only be seen whenever we have a basic plan that is constantly compared to our real actions.
\c. \textsc{EN}: The women wish that the \{painters, victims\} walk consciously, and every 2CV or Dyane can basically be used as a donor.

Interestingly, the models sometimes also changed the order in the relative clause when a scope change was not possible, for instance when the second conjunct was a question, or the verb in the first sentence did not allow to take scope over the second conjunct without the presence of the word \exa{that}.
See Example~\ref{ex:vaders_president}. We underline the incorrect part of the translation, here and in erroneous examples that follow.

\ex. \label{ex:vaders_president}
\a. \textsc{EN}: The victim observes the \{leader, king\}, and the fathers carefully avoid the president.
\b. \textsc{NL}: Het slachtoffer observeert de leider en de vaders \underline{de president zorgvuldig vermijden}.
\c. \textsc{NL}: Het slachtoffer observeert de koning en de vaders vermijden voorzichtig de president.

These examples indicate that the interpretation of scope change might not be applicable here and that instead, the model is applying some heuristic where particular words trigger a relative clause order.

\subsubsection{Target errors}
In the category `target errors', some of the errors can be easily traced to individual words, whereas others indicate overall misinterpretations of the input.

\paragraph{Single word errors}
Errors that consist of single words are caused by words that are either missing, wrongly translated or untranslated.
Changes due to \textbf{missing words} can be very minor but nevertheless render one of the sentences ungrammatical (e.g.\ \exa{De tante achter de truck bewonderde de directeur}, correct, vs \exa{De tante achter de truck bewonderde directeur}, incorrect),
or yield grammatical sentences that have a slightly different meaning (e.g.\ \exa{de arts die yoghurt eet} vs \exa{de arts die \emph{de} yoghurt eet}).
Missing words can also render translations both ungrammatical and semantically incorrect, which occurred mostly in case of missing nouns or verbs (e.g. \exa{de bakker die ons herkent, merkt de koning op}, correct, vs \exa{de bakker die ons de koning herkent}, incorrect).

We also encountered pairs where one translation contained \textbf{untranslated source words}.
This happened with some of the words in our synthetic templates (e.g. \exa{ooms}/\exa{uncles}, \exa{butchers}/\exa{slagers}) but also with words from the natural sentences (e.g.\ \exa{extrusion}/\exa{extrusie}, \exa{soils}/\exa{bodem}).
These cases mark examples where local processing would have been helpful to the model: as evidenced by the alternative translation in the pair, the model does have access to the correct translation.

Thirdly, we observed cases of \textbf{mistranslated words}, where words unrelated to the change locus received a wrong translation in one of the two sentences but a correct one in the other, for example:
	\exa{poets} being translated as \exa{dichters} (correct) vs \exa{de potten} (incorrect), \exa{general} as \exa{generaal} (correct) vs \exa{wandeling} (incorrect), or \exa{productform} as \exa{productvorm} (correct) vs \exa{productformulier} (incorrect).

\paragraph{Multi-word errors} Other types of errors are less easily located to individual words but indicate an overall misinterpretation of the input, such as the \textbf{change in the tense} as displayed in Example~\ref{ex:musicians},
and the \textbf{change in agreement} displayed in Example~\ref{ex:begrijpen}.
In these particular cases, the source of confusion is explainable: in the first case, the model is combining a present tense verb with a word-order that does not support that, even though such a word order does exist (\exa{in het najaar van 2005 \dots en komen er al snel een paar \dots}).
In the second case, \exa{begrijpen} should agree with \exa{schilder} but instead agrees with the word \exa{doctors}, much earlier in the sentence.
In both of these cases, a more locally compositional approach to translating would have yielded correct translations.

\ex. \label{ex:musicians}
\a. \textsc{EN}: (\dots) and in autumn 2005, five musicians join their forces and soon a couple of potential songs came into being in the rehearsal room.
\b. \textsc{NL}: (\dots) in het najaar van 2005 voegen vijf muzikanten zich bij hun krachten en al snel kwamen er een paar potentiële nummers in de oefenruimte.
\c. \textsc{NL}: (\dots) in het najaar van 2005 bundelen vijf muzikanten hun krachten en al snel \underline{komen} er een paar potentiële nummers tot stand in de oefenruimte.

\ex. \label{ex:begrijpen}
\a. \textsc{EN}: The doctors that laugh admire the \{president, baker\}, the painter that admires her understands the king.
\b. \textsc{NL}: (\dots) de schilder die haar bewondert, \underline{begrijpen} de koning.
\c. \textsc{NL}: (\dots) de schilder die haar bewondert begrijpt de koning.

Finally, we would like to point out an error type that relates to the \textbf{semantic role assigned to agents}, and brings about a lot of other changes in the process.
For instance, in Example~\ref{ex:father}, \exa{the fathers} is removed from the main clause and moved into the relative clause, leaving the main clause without its direct object.

\ex. \label{ex:father}
\a. \textsc{EN}: The group of painters behind the truck forgets the \{president, friend\} and an article about the previous EESC Opinion on alcohol related harm, which looked at f, is read by the fathers
\b. \textsc{NL}: (\dots) en een artikel over het eerdere advies van het EESC over alcoholgerelateerde schade, \underline{die door de vaders wordt onderzocht, wordt gelezen}.
\c. \textsc{NL}: (\dots) en een artikel over het eerdere advies van het EESC over alcoholgerelateerde schade, die naar f uitkeek, wordt door de vaders gelezen.

\subsubsection{Formatting}

We marked inconsistencies as formatting changes if they were related to punctuation, capitalisation, hyphenation or differences in usage of spaces. 
In most cases, those cases were caused by commas: in one translation, a relative clause or two conjuncts were separated by a comma, whereas in the other translation the comma was left out.
In the cases that were caused by spaces (\exa{tumormassa} vs \exa{tumor massa}), there is a slight difference in correctness: in Dutch, compound nouns are not separated by spaces. 
Given how minor these mistakes are, we did not mark them as errors.
Example~\ref{ex:begrijpen} above provides an example for inconsistent usage of commas.
Formatting changes are far from the most frequent but they do become more prominent in models trained on larger training corpora.

\subsubsection{Inconsistencies in synonym translations}

The synonym errors are subdivided into cases where synonyms are simply translated differently (we observed this mostly for the models with larger training set sizes), cases where both translations were incorrect, cases in which only one translation is wrong, and cases in which one synonym was not translated but directly copied from the source.
Sometimes, the changes were quite peculiar, to give some examples from our natural corpus:

\ex.
\a. \textsc{EN}: The child admires the king that eats the \{doughnut, donut\}.
\b. \textsc{NL}: Het kind bewondert de koning die de donut eet.
\c. \textsc{NL}: Het kind bewondert de koning die de \underline{ezel} eet.

\ex.
\a. \textsc{EN}: - Yeah, a barbecue sauce \{moustache, mustache\} contest.
\b. \textsc{NL}: - Ja, een barbecue \underline{[missing `sauce']} met snor.
\c. \textsc{NL}: - Ja, een barbeceu saus snor wedstrijd.

How often each of these errors occur depends on the synonym.
Where some synonyms are more prone to being untranslated (like \exa{ladybird} and \exa{flautist}), some simply received many different correct translations (like \exa{shopping trolley}) yet others received errors very specific to the synonym (like \exa{eggplant} being translated as \exa{egg}$+$\exa{plant}, an interesting case because it reflects processing that is too local).
It should be noted that for all synonyms -- apart from the model with the small training dataset that cannot translate \exa{flautist} and \exa{ladybug} -- we have observed correct translations, indicating that the models did in fact acquire their meaning.

Further, it should be noted that while our substitutivity experiment provides insight into how the model copes with individual synonyms, the majority of the inconsistencies observed were still common target errors, rephrasings, changes in formatting or the result of source-side ambiguities. 
It is vital here to stress that the types of rephrasings, however, did not appear related to the writing style of the sentence. For instance, considering that the synonym changes were related to British and American spelling, and occassionally changed the tone of the sentence (e.g.\ \exa{aeroplane} could be considered more archaic compared to \exa{airplane}), one could anticipate changes in word choice in Dutch reflecting this change of style. However, the inconsistencies were virtually indistinguishable from those annotated for systematicity.